\documentclass[12pt]{article}
\usepackage{amsmath}
\usepackage[round]{natbib}
\usepackage{amssymb,color}
\usepackage{appendix}
\usepackage{amsthm}
\usepackage{hyperref}
\usepackage{enumitem}
\usepackage{dsfont}
\usepackage{mathtools} 
\mathtoolsset{showonlyrefs=true}  

\usepackage{caption}

\usepackage{algorithm}
\usepackage{algorithmicx, algpseudocode}

\usepackage{booktabs}

\usepackage[margin=0.9in]{geometry}

\numberwithin{equation}{section}
\usepackage{comment}

\usepackage{titling}
\newtheorem{theorem}{Theorem}[section]
\newtheorem{corollary}[theorem]{Corollary}
\newtheorem{definition}{Definition}[section]

\newtheorem{proposition}[theorem]{Proposition}

\newtheorem{lemma}[theorem]{Lemma}

\begin{document}

	\author{Wonjae Lee\thanks{Department of Mathematical Sciences, Seoul National University, Seoul 08826, South Korea. phwjlee117@gmail.com},  Taeyoung Kim \thanks{School of Computational Sciences, Korea Institute for Advanced Study, Seoul 02455, South Korea. taeyoungkim@kias.re.kr},  Hyungbin Park\thanks{Research Institute of Mathematics \& Department of Mathematical Sciences, Seoul National University, Seoul 08826, South Korea. hyungbin@snu.ac.kr, hyungbin2015@gmail.com}     
	}

\title{
	Fourier Neural Operators for Non-Markovian Processes: Approximation Theorems and Experiments
}

	\maketitle

\begin{abstract}
	This paper introduces an operator-based neural network, the mirror-padded Fourier neural operator (MFNO), designed to learn the dynamics of stochastic systems. MFNO extends the standard Fourier neural operator (FNO) by incorporating mirror padding, enabling it to handle non-periodic inputs. We rigorously prove that MFNOs can approximate solutions of path-dependent stochastic differential equations and Lipschitz transformations of fractional Brownian motions to an arbitrary degree of accuracy. Our theoretical analysis builds on Wong--Zakai type theorems and various approximation techniques. Empirically, the MFNO exhibits strong resolution generalization--a property rarely seen in standard architectures such as LSTMs, TCNs, and DeepONet. Furthermore, our model achieves performance that is comparable or superior to these baselines while offering significantly faster sample path generation than classical numerical schemes.\\

\end{abstract}

\section{Introduction}

\subsection{Overview}

Stochastic processes are foundational tools for modeling systems governed by randomness. These processes provide a mathematical framework for describing the temporal evolution of uncertain phenomena, with wide-ranging applications in finance, physics, biology, and engineering. Stochastic differential equations (SDEs), typically driven by Brownian motion, form the core analytical framework for modeling stochastic processes. For example, they are used to model asset prices and interest rates in finance, particle diffusion and thermal fluctuations in physics, population dynamics and neural activity in biology, and signal processing and control systems in engineering. Recent advances in machine learning have introduced novel perspectives and powerful methodologies for analyzing SDEs.

This paper develops a novel operator-based neural network approach for SDEs. We utilize Fourier neural operator (FNOs) to learn the solution operator associated with SDEs. Consider the SDE
\begin{equation}
\label{eqn:SDEs} 
dX(t) = b(t, X(t))\,dt + \sigma(t, X(t))\,dB(t)\,, \; X(0) = \xi\,,
\end{equation}
which defines an operator
$$
X = F(\xi, B),
$$
where $F$ maps the initial condition $\xi\in\mathbb{R}^m$ and the Brownian path $B\in C([0,T], \mathbb{R}^\ell)$ to the solution path $X\in C([0,T], \mathbb{R}^m)$. We regard $\xi$ as a constant path and extend the operator to
$$
F: C([0,T], \mathbb{R}^{m+\ell}) \to C([0,T], \mathbb{R}^m),
$$
which maps the paired path $(\xi,B)$ to the solution $X$. The core idea of our work is to approximate this solution operator $F$ using an FNO. This operator-learning approach is applicable to a broad class of SDEs and offers significant modeling flexibility. By operating directly on function spaces, the FNO can capture complex temporal dependencies more effectively. Moreover, its non-local kernel representation renders it particularly well-suited for learning the global dynamics of stochastic systems.

However, directly approximating the operator $F$ with an FNO is not straightforward and presents several challenges. A primary challenge is that the operator $F$ is merely measurable, not continuous. Although FNOs are well-suited for approximating continuous operators on Sobolev spaces (\cite{Kovachki:21}), the feasibility of utilizing them to effectively approximate measurable operators acting on the space of continuous paths remains unclear. To overcome this, we employ the Wong--Zakai approximation, where the Brownian motion is replaced by its piecewise linear interpolation. This approximation bridges the gap between measurable operators on the space of continuous functions and continuous operators on Sobolev spaces.

A second challenge arises from the non-periodic nature of Brownian paths on a finite interval $[0, T]$, which conflicts with the FNO's inherent assumption of periodicity. To resolve this, we introduce the mirror-padded FNO (MFNO), an architecture where the input path is symmetrically extended by reflecting it about the midpoint $T$. This construction produces a continuous and periodic function over the extended interval $[0, 2T]$, satisfying the theoretical requirements for the FNO to operate effectively.

\subsection{Main contributions}

The contributions of this study are summarized as follows.
First, we introduce a novel neural operator framework specifically designed for learning a broad class of SDEs, particularly those with non-Markovian dynamics. To our knowledge, this is the first study to adapt FNOs for learning SDE solution operators in a stochastic setting. Our approach effectively captures the temporal and stochastic structures inherent in SDEs by leveraging the global representation capabilities of FNOs. In particular, our method performs well when learning a wide range of stochastic processes, including path-dependent SDEs and systems driven by fractional Brownian motion (fBM). 
This framework opens up new possibilities for the efficient and broadly applicable learning of stochastic systems using operator-learning methods.

Second, we establish the first rigorous approximation theorems for learning path-dependent SDEs and Lipschitz transformations of fBM using an FNO-based architecture. While numerous experimental works have explored neural networks for approximating stochastic processes, theoretical justification has remained scarce. We prove that MFNOs, when given a linearly interpolated Brownian motion as input, can approximate the solutions of path-dependent SDEs and Lipschitz transformations of fBM with any desired accuracy. These results fill a significant theoretical gap and formally demonstrate the capability of FNO-based architectures for learning complex stochastic systems.

Third, our empirical results demonstrate strong performance, offering superior resolution generalization and computational efficiency compared to established methods. Resolution generalization is a neural network's ability to generate outputs at higher temporal resolutions than seen during training without loss in accuracy. Existing architectures such as LSTMs, TCNs, neural SDEs, and DeepONets are typically tied to the time grid on which they were trained and therefore rarely exhibit this property. In contrast, MFNOs excel in this regard, indicating that they learn a resolution-invariant operator capable of effectively interpolating from coarse temporal inputs. Furthermore, our method yields improved computational efficiency. By leveraging the FNO architecture, sample generation scales as $O(n \log n)$, a significant improvement over the $O(n^2)$ complexity of the Euler scheme commonly used for generic path-dependent SDEs. This theoretical advantage is confirmed by our experimental results, which show substantial practical gains in efficiency.

\subsection{Related literature}

\paragraph{Non-Markovian dynamics}
Path-dependent SDEs represent a non-Markovian extension of standard SDEs. Functional Itô calculus \citep{D:09} and its subsequent extensions \citep{CF:10} provide a rigorous analytical framework for such systems. Building on this foundation, \cite{CL:16} generalized the Euler scheme to handle path-dependent SDEs. These developments have enabled a range of applications, including stochastic control \citep{S:19} and option pricing \citep{LLP:22}. fBM introduces long-range dependence via correlated increments and has found applications in various domains, including finance and network modeling \citep{RS:13, GJR:18, No:95}. Simulation methods for Gaussian processes, including fBM, are discussed in \cite{H:84} and \citep{AG:07}.

\paragraph{Neural SDE}
Neural SDEs were first introduced in the seminal work by \cite{tzen2019neural}. Since then, \cite{KFLOL:21} has interpreted neural SDEs as operators between function spaces, an approach conceptually dual to ours. Neural SDEs have also been extended to incorporate fractional white noise in \cite{TNTC:22}. Although neural SDEs exhibit mesh invariance and can be extended to capture certain non-Markovian processes, they do not cover the full class of path-dependent SDEs or fBMs addressed in our work.

\paragraph{Neural operators}  
\cite{Li:21} studied FNOs for parametric PDEs and demonstrated their strong performance in learning global dynamics, including zero-shot super-resolution capabilities. \cite{Kovachki:21} established the universality of FNOs by proving that they can approximate any continuous operator between Sobolev spaces to a desired accuracy. \cite{HMCGWC:22} and \cite{LDPH:24} applied FNOs to learn stochastic partial differential equations. Beyond Fourier-based architectures, several alternative neural operator frameworks have been proposed, including graph neural operators \citep{LKALBSA:20} and Laplace neural operators \citep{CGK:24}. DeepOnets have also been explored for learning solutions to SDEs in \cite{LL:23} and \cite{EM:25}. Most recently, \cite{SRAA:25} introduced the flow matching method for neural operators.

\paragraph{Time-series neural networks} 
\cite{HJ:97} introduced long short-term memory (LSTM) networks to capture long-range dependencies in sequential data using gating mechanisms to control information flow. \cite{Lea:16} proposed temporal convolutional networks (TCNs), which employ dilated causal convolutions to model long-term dependencies without recurrence. Although LSTMs and TCNs can extrapolate learned dynamics and handle inputs of arbitrary length, they operate on a fixed grid resolution and lack the ability to interpolate or generalize across multiple temporal resolutions.

$ $

The remainder of this paper is organized as follows. Section \ref{sec:model} reviews the FNO and its universal approximation theorems, then introduces our key modifications: mirror padding and the use of linearly interpolated Brownian motion. In Section \ref{sec:conv}, we present and prove our approximation theorems for path-dependent SDEs and fBMs. Section \ref{sec:exper} presents experimental results, comparing our MFNO approach against zero-padded FNOs, FNOs without padding, LSTMs, TCNs, and DeepONets in terms of accuracy, speed, and resolution generalization. Finally, Section \ref{sec:con} concludes the paper.

\section{Model architecture}\label{sec:model}

This section reviews the architecture and universal approximation property of FNOs and then introduces linearly interpolated Brownian motions and our proposed mirror-padded FNOs.

\subsection{Fourier neural operator}
\label{sec:FNO}
\noindent

\begin{figure}[htp!]
	\centering
	\includegraphics[height=6.0cm]{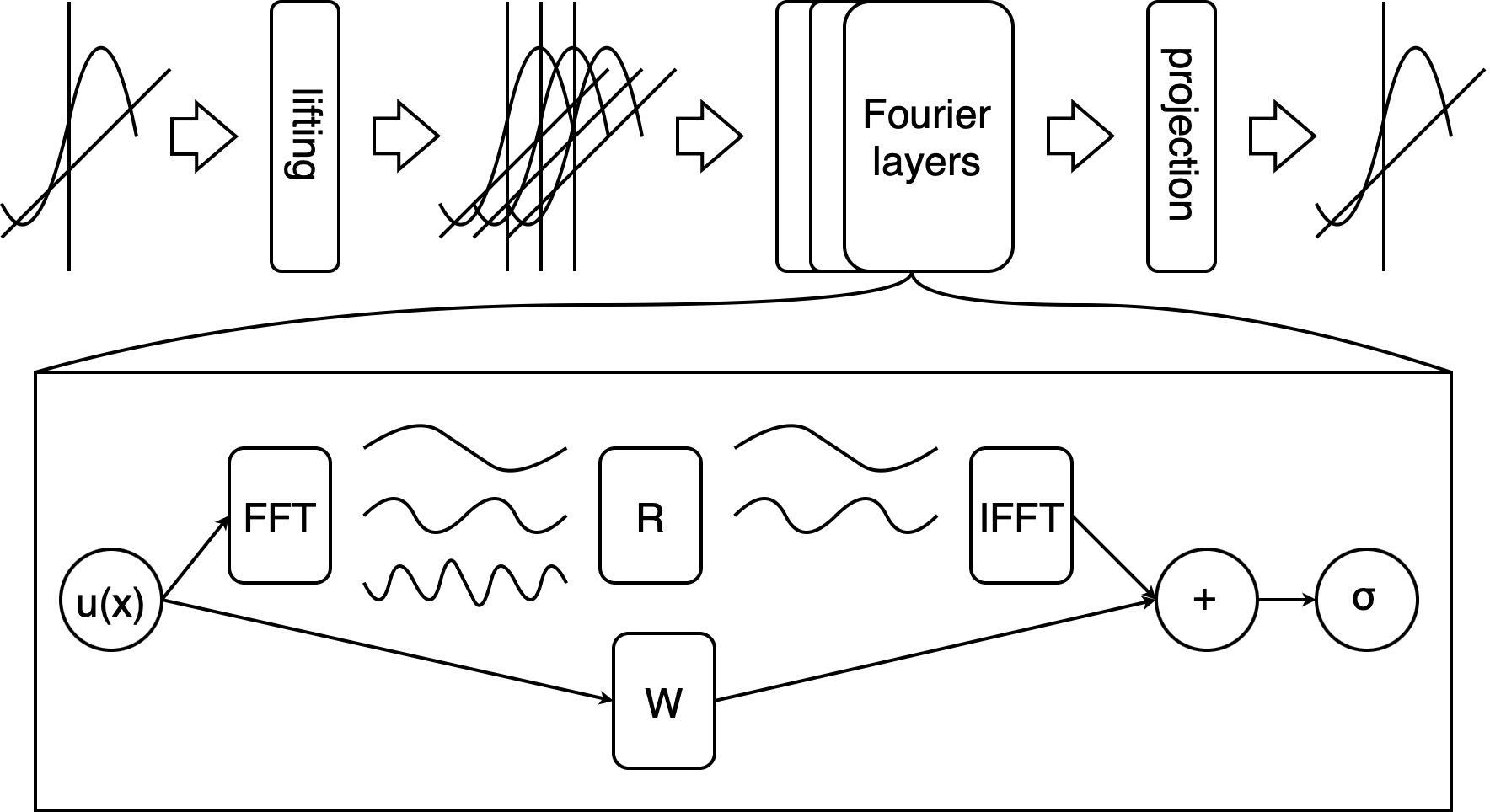}
	
	\caption{Schematic of a 1D FNO}\label{1dFNO}
\end{figure}

We begin with a review of the basic concepts of Fourier neural operators and their universal approximation theorems, closely following the framework developed by Kovachki \textit{et al.}~\cite{Kovachki:21}. Let $\mathcal{A}(\mathcal{D},\mathbb{R}^{d_{a}})$ and $\mathcal{U}(\mathcal{D},\mathbb{R}^{d_{u}})$ be suitable Banach spaces of $\mathbb{R}^{d_{a}}$-valued and $\mathbb{R}^{d_{u}}$-valued functions, respectively, on a subset $\mathcal{D}\subset \mathbb{R}^{d}$. A typical neural operator $\mathcal{N}:\mathcal{A}(\mathcal{D},\mathbb{R}^{d_{a}})\to \mathcal{U}(\mathcal{D},\mathbb{R}^{d_{u}})$ has the following structure:
\[
\mathcal{N}=\mathcal{Q}\circ\mathcal{L}_{L}\circ\mathcal{L}_{L-1}\cdots\circ\mathcal{L}_{1}\circ\mathcal{R}\,.
\]
Here, $\mathcal{R}:\mathcal{A}(\mathcal{D},\mathbb{R}^{d_{a}})\to\mathcal{U}(\mathcal{D},\mathbb{R}^{d_{v}})$ and $\mathcal{Q}:\mathcal{U}(\mathcal{D},\mathbb{R}^{d_{v}})\to \mathcal{U}(\mathcal{D},\mathbb{R}^{d_{u}})$ are the lifting and projection layers, respectively. The lifting layer $\mathcal{R}$ elevates the input data to a higher-dimensional feature space, while the projection layer $\mathcal{Q}$ maps the features back to the target dimension. Specifically, they often take the form
\begin{align}
\label{lift}
\mathcal{R}(a)(x)=Ra(x)\,, \quad R\in\mathbb{R}^{d_{v}\times d_{a}}\,,\\
\label{project}
\mathcal{Q}(v)(x)=Qv(x)\,, \quad Q\in\mathbb{R}^{d_{u}\times d_{v}}\,.
\end{align}
Each $\mathcal{L}_{l}:\mathcal{U}(\mathcal{D},\mathbb{R}^{d_{v}})\to \mathcal{U}(\mathcal{D},\mathbb{R}^{d_{v}})$, for $l=1,2,\cdots, L$, is a non-linear layer comprising a kernel integration and an affine pointwise mapping. It specifically has the form
\begin{equation}
\label{Flayer}
\mathcal{L}_{l}(v)(x)=\sigma\left( W_{l}v(x)+b_{l}(x)+\int_{D}\kappa_{\theta_{l}}\left(x,y,a(x),a(y)\right)v(y)dy\right).
\end{equation}
Here, $a\in\mathcal{A}(\mathcal{D},\mathbb{R}^{d_{a}})$ is the initial input to the neural operator, $W_{l}\in\mathbb{R}^{d_{v}\times d_{v}}$ is a weight matrix, $b_{l}\in\mathcal{U}(\mathcal{D},\mathbb{R}^{d_{v}})$ is a bias term, the kernel $\kappa_{\theta_{l}}:\mathcal{D}\times \mathcal{D}\times \mathbb{R}^{d_{a}}\times\mathbb{R}^{d_{a}}\to\mathbb{R}^{d_v\times d_v} $ is a neural network parameterized by $\theta_{l}$, and $\sigma:\mathbb{R}\to \mathbb{R}$ is a non-polynomial, Lipschitz continuous, and $C^{3}$ activation function applied component-wise.

A Fourier neural operator (FNO) is constructed by considering a periodic domain $\mathcal{D}=\mathbb{T}^{d}=[0,2\pi]^{d}/\sim$ and a kernel of the form 
\[\kappa_{\theta_{l}}(x,y,a(x),a(y))=\kappa_{\theta_{l}}(x-y)\,,\,x,y\in \mathbb{T}^{d}\,.\]
Let $\mathcal{F}$ denote the Fourier transform. The Fourier transform of the kernel is
\[P_{\theta_{l}}(k)=\mathcal{F}(\kappa_{\theta_l})(k)=\frac{1}{(2\pi)^{d}}\int_{\mathbb{T}^{d}} \kappa_{\theta_{l}}(x) e^{-i\langle k,x\rangle}dx\in\mathbb{C}^{d_{v}\times d_{v}}, \quad k\in\mathbb{Z}^{d}.\]
By the convolution theorem, the kernel integration in \eqref{Flayer} can be expressed as a product in the Fourier domain:
\[ \int_{D}\kappa_{\theta_{l}}(x-y)v(y)dy = \mathcal{F}^{-1}\left(P_{\theta_{l}}\cdot \mathcal{F}(v)\right)(x).\]
Thus, the non-linear layers of an FNO, known as Fourier layers, take the following form:  
\begin{equation}
\label{flayer}
\mathcal{L}_{l}(v)(x)=\sigma\left(W_{l}v(x)+b_l(x)+\mathcal{F}^{-1}\left(P_{\theta_{l}}\cdot \mathcal{F}(v)\right)(x)\right).
\end{equation}
Notably, instead of parameterizing the kernel functions $\kappa_{\theta_{l}}$, we can directly parameterize $\mathcal{L}_{l}$ with the Fourier weights $P_{\theta_{l}}(k)$ under the constraint $P_{\theta_{l}}(-k)=P_{\theta_{l}}(k)^{\dagger}$.

We now formally define the FNO and state its universal approximation property. Let $H^{s}(\mathcal{D},\mathbb{R}^{m})$ be the Sobolev space of functions from $\mathcal{D}$ to $\mathbb{R}^{m}$ with smoothness $s\ge 0$, equipped with the norm $\|\cdot\|_{H^{s}}$.
\begin{definition}[FNO]
	Let $s,s'\ge 0$ and $d,d_a,d_u\in\mathbb{N}.$ 
	An FNO is a map  $\mathcal{N}:H^{s}(\mathbb{T}^{d},\mathbb{R}^{d_{a}})\to H^{s'}(\mathbb{T}^{d},\mathbb{R}^{d_{u}})$ given as
	\[\mathcal{N}=\mathcal{Q}\circ\mathcal{L}_{L}\circ\mathcal{L}_{L-1}\cdots\circ\mathcal{L}_{1}\circ\mathcal{R},\]
	where $\mathcal{R}$, $\mathcal{Q}$, and $\mathcal{L}_{1},\cdots,\mathcal{L}_L$ are of the form in \eqref{lift}, \eqref{project}, and \eqref{flayer}, respectively.
\end{definition}

\begin{theorem}[Universal approximation for FNOs]
	\label{uatfno}
	Let $s,s'\ge 0$ and $d,d_a,d_u\in\mathbb{N}.$ Suppose $\mathcal{G}:H^{s}(\mathbb{T}^{d};\mathbb{R}^{d_{a}}) \to H^{s'}(\mathbb{T}^{d};\mathbb{R}^{d_{u}})$ is a continuous operator and $K\subset H^{s}(\mathbb{T}^{d};\mathbb{R}^{d_{a}})$ is a compact subset. Then, for any $\epsilon>0$, there exists an FNO $\mathcal{N}:H^{s}(\mathbb{T}^{d};\mathbb{R}^{d_{a}})\to H^{s'}(\mathbb{T}^{d};\mathbb{R}^{d_{u}})$ such that
	\[\sup_{a\in K} \|\mathcal{G}(a)-\mathcal{N}(a)\|_{H^{s'}} \le \epsilon.\]
\end{theorem}

For practical implementation, we introduce the $\Psi$-FNO, a discretized version of the FNO. A true FNO cannot be directly implemented on a computer, as it requires computing an infinite number of Fourier weights $P_{\theta_{l}}(k)$ and Fourier coefficients $\mathcal{F}(v)(k)$ for all $k \in \mathbb{Z}^{d}$. In practice, a frequency cutoff $W$, referred to as the width of the FNO, is introduced. Specifically, the Fourier weights are truncated by setting $P_{\theta_{l}}(k)=0$ whenever $|k|_{\infty}:=\max_{1\le i\le d}|k_i|>W$. Additionally, as computers can only handle a finite number of input points, we fix a regular grid $\mathcal{J}_{N}:=\{2\pi j/(2N+1)\}_{j\in \mathbb{Z}^{d}}$ on the torus $\mathbb{T}^{d}$ for some $N\in \mathbb{N}$. The input function $v$ is then projected into a trigonometric polynomial of degree $N$ before each Fourier layer. This discretized version of the FNO is referred to as the $\Psi$-FNO.
Let $C_{N}(\mathbb{T}^{d}, \mathbb{R}^{d_{u}})$ denote the space of $\mathbb{R}^{d_{u}}$-valued trigonometric polynomials of order $N$ on the torus $\mathbb{T}^{d}$. We define the pseudo-spectral Fourier projection operator $\mathcal{I}_{N} : C(\mathbb{T}^{d}, \mathbb{R}^{d_{v}}) \to C_N(\mathbb{T}^{d}, \mathbb{R}^{d_{v}})$ as the projection onto trigonometric polynomials of order $N$. That is, for any $v \in C(\mathbb{T}^{d}, \mathbb{R}^{d_{v}})$, the projection $\mathcal{I}_{N}(v) \in C_N(\mathbb{T}^{d}, \mathbb{R}^{d_{v}})$ is the unique trigonometric polynomial of order $N$ that satisfies
$$
\mathcal{I}_{N}(v)(x) = v(x), \quad x \in \mathcal{J}_{N}.
$$
The Fourier coefficients of $\mathcal{I}_{N}(v)$ can be efficiently computed by applying the discrete Fourier transform (DFT) to the sequence $(v(x))_{x\in\mathcal{J}_{N}}$.

\begin{definition}[$\Psi$-FNO]
	Let $d,d_a,d_u\in\mathbb{N}$ and $N\in \mathbb{N}$, and let $\mathcal{A}(\mathbb{T}^{d},\mathbb{R}^{d_{a}})$ be a Banach space of $\mathbb{R}^{d_{a}}$-valued continuous functions on $\mathbb{T}^{d}$. A $\Psi$-FNO with order $N$ is a map $\mathcal{N}:\mathcal{A}(\mathbb{T}^{d},\mathbb{R}^{d_{a}})\to C_N(\mathbb{T}^{d},\mathbb{R}^{d_{u}})$ given as
	\begin{equation} 
	\mathcal{N}=\mathcal{Q}\circ\mathcal{I}_{N}\circ\mathcal{L}_{L}\circ\mathcal{I}_{N}\circ\mathcal{L}_{L-1}\cdots\circ\mathcal{L}_{1}\circ\mathcal{I}_{N}\circ\mathcal{R},
	\end{equation}
	where $\mathcal{R}: \mathcal{A}(\mathbb{T}^{d},\mathbb{R}^{d_{a}} ) \to C (\mathbb{T}^{d},\mathbb{R}^{d_v})$ is a lifting layer, $\mathcal{Q}:C_N(\mathbb{T}^{d},\mathbb{R}^{d_{v}})\to C_N(\mathbb{T}^{d},\mathbb{R}^{d_{u}})$ is a projection layer, and $\mathcal{L}_{1},\cdots,\mathcal{L}_L:C_N(\mathbb{T}^{d},\mathbb{R}^{d_{v}})\to C(\mathbb{T}^{d},\mathbb{R}^{d_{v}})$ are non-linear layers of the form in \eqref{lift}), \eqref{project}, and \eqref{flayer}, respectively.
\end{definition}

We now state the universal approximation theorem for $\Psi$-FNOs in a Sobolev space setting. By the Sobolev embedding theorem, $H^{s}(\mathbb{T}^{d}, \mathbb{R}^{m})$ is compactly embedded in $C(\mathbb{T}^{d}, \mathbb{R}^{m})$ when $s > d/2$. Throughout this paper, we identify $H^{s}(\mathbb{T}^{d}, \mathbb{R}^{m})$ with its image in $C(\mathbb{T}^{d}, \mathbb{R}^{m})$ via this embedding and regard it as a subset of $C(\mathbb{T}^{d}, \mathbb{R}^{m})$.
\begin{theorem}[Universal approximation for $\Psi$-FNOs]
	\label{uatpfno}
	Let $s>d/2$ and $s'\ge 0$. Suppose $\mathcal{G}:H^{s}(\mathbb{T}^{d},\mathbb{R}^{d_{a}})\to H^{s'}(\mathbb{T}^{d},\mathbb{R}^{d_{u}})$ is a continuous operator and let $K\subset H^{s}(\mathbb{T}^{d},\mathbb{R}^{d_{a}})$ be a compact subset. Then, for any $\epsilon>0$, there exist $N\in\mathbb{N}$ and a $\Psi$-FNO $\mathcal{N}:H^{s}(\mathbb{T}^{d},\mathbb{R}^{d_{a}})\to C_N(\mathbb{T}^{d},\mathbb{R}^{d_{u}})$ with order $N$ such that
	\[\sup_{a\in K} \|\mathcal{G}(a)-\mathcal{N}(a)\|_{H^{s'}}\le \epsilon.\]
\end{theorem}


\subsection{Linearly interpolated Brownian motions and mirror-padded FNOs}

We begin by defining linearly interpolated Brownian motions. This construction is key to enabling the use of Wong--Zakai-type approximations and related theoretical results from \cite{DU:99} as inputs to our model.
\begin{definition}
	Let $B$ be an $\ell$-dimensional Brownian motion, and let $\pi_{n}:=\{0=t^{n}_{0}<t_{1}^n<\cdots<t_{N_{n}^n}=T\}\, (n\in\mathbb{N})$ be an increasing sequence of uniform partitions of $[0,T]$. Then, the non-adapted piecewise linear interpolation of $B$ with respect to $\pi_{n}$ is a process defined by
	\[B^{n}(t)=B(t^{n}_{i})+\frac{B(t^{n}_{i+1})-B(t^{n}_{i})}{t^{n}_{i+1}-t^{n}_{i}}(t-t^{n}_{i})\,,\; t\in [t^{n}_{i},t^{n}_{i+1})\,.\]
\end{definition}

Notably, the finite set of values $(0, B^{n}(t_{1}), \ldots, B^{n}(t_{N_{n}}))$ completely determines the entire path $B^n$. We use these values as inputs to our model to represent the sample path $B^n$. Moreover, for each $n \in \mathbb{N}$, the set of sample paths of $B^n$ lies within a finite-dimensional subspace of $H^{1}([0,T], \mathbb{R}^{d})$, a fact that will be useful in our convergence analysis. A computer can only process a finite number of values, so in practice, a finite set of points from a Brownian motion sample path is used as input, rather than the entire path, which consists of uncountably many points. A key insight is that these input points can be interpreted as lying on the sample path of either a linearly interpolated Brownian motion or a true Brownian motion. While this distinction does not affect the actual computation during the feedforward process, it plays a crucial role in the mathematical analysis presented in Section \ref{sec:conv}.

The following lemma establishes a probabilistic property of the non-adapted piecewise linear interpolation $B^n$, which will be used to prove our main approximation results.

\begin{lemma}
	\label{lemmaPB}
	For any $0<\epsilon<1$ and $M\in\mathbb{N}$, there exists a constant $R_{M,\epsilon}>0$ such that 
	\[\mathbb{P}(\|B^{n}\|_{H^{1}}\le R_{M,\epsilon})\ge 1-\epsilon\]
	for all $n\le M$.
\end{lemma}
\noindent
\begin{proof}
	For simplicity, we assume $B$ is one-dimensional; the proof for the multi-dimensional case is analogous. Let $\pi_n:=\{0=t^{n}_{0}<t^{n}_{1}<\cdots<t^{n}_{N_{n}}=T\}$ be the sequence of partitions of $[0,T]$. We first estimate the $L_2$-norm of $(B^n)'$, the weak derivative of $B^n.$ Observe that 
	\begin{align*}
	\left\|(B^{n})'\right\|_{L^{2}}^{2}
	&= \int_{0}^{T} \left|(B^{n})'(t)\right|^{2}\,dt \\
	&= \sum_{i=0}^{N_{n}-1} \int_{t^{n}_{i}}^{t^{n}_{i+1}}
	\left| \frac{B(t^{n}_{i+1}) - B(t^{n}_{i})}{t^{n}_{i+1} - t^{n}_{i}} \right|^{2} dt \\
	&= \sum_{i=0}^{N_{n}-1}
	\left( \frac{B(t^{n}_{i+1}) - B(t^{n}_{i})}{\sqrt{t^{n}_{i+1} - t^{n}_{i}}} \right)^{2}.
	\end{align*}
	Define $X_{i}:=\frac{B(t^{n}_{i+1})-B(t^{n}_{i})}{\sqrt{t^{n}_{i+1}-t^{n}_{i}}}$ for each $i\in\{0,\cdots, N_{n}-1\}$. Then, 
	\[X_{i}\sim  N(0,1)\quad \text{i.i.d. for}\; i=0,\cdots, N_{n-1}.\]
	Thus,  $\|(B^{n})'\|^{2}_{L^{2}}=\sum^{N_{n}-1}_{i=0}X^{2}_{i}$ follows the $\chi^{2}$-distribution with $N_{n}$ degrees of freedom. The concentration inequality for $\chi^{2}$-distributions yields
	\[\mathbb{P}\left(\|(B^{n})'\|^{2}_{L^{2}}> N_{n}+2\sqrt{N_{n}x}+2x\right)\le e^{-x}\] for all $x\ge0.$
	Since $N_{M}\ge N_{n}$ for all $n\le M$, we obtain
	\begin{equation}
	\label{5}
	\mathbb{P}\left(\|(B^{n})'\|^{2}_{L^{2}}> N_{M}+2\sqrt{N_{M}x}+2x\right)\le e^{-x}
	\end{equation}
	for $x\ge0$.

	Observe that
	$\|B^{n}\|^{2}_{H^{1}}= \|B^{n}\|_{L^{2}}^2+\|(B^{n})'\|_{L^{2}}^2 \le (1+T^{2})\|(B^{n})'\|^{2}_{L^{2}}$
	by a Poincare-type inequality, $\|B^{n}\|_{L^{2}}\le T\|(B^{n})'\|_{L^{2}}$. Thus, for any $R>0$, we have
	\begin{equation}
	\label{6}
	\mathbb{P}\left(\|B^{n}\|^{2}_{H^{1}}> (1+T^{2})R\right)\le\mathbb{P}\left(\|(B^{n})'\|^{2}_{L^{2}}> R\right).
	\end{equation}
	Setting $x=\ln(1/\epsilon)$ in \eqref{5} and combining it with \eqref{6}, we obtain the desired inequality
	\[\mathbb{P}(\|B^{n}\|^{2}_{H^{1}}\le R_{M,\epsilon})\ge 1- \epsilon\]
	for \[R_{M,\epsilon}:=(1+T^{2})\left(N_{M}+2\sqrt{N_{M}\ln(1/\epsilon)}+2\ln(1/\epsilon)\right).\]
	This completes the proof.
	\qed
\end{proof}

Next, we introduce mirror-padded FNOs. Without loss of generality, we assume $T = \pi$ throughout the remainder of this section and Section \ref{sec:conv}. As established in Theorem \ref{uatpfno}, the universal approximation guarantee for $\Psi$-FNOs holds only for periodic inputs. A sample path of $B^n$, however, generally does not satisfy periodic boundary conditions, precluding a direct application of Theorem \ref{uatpfno}. A standard practice to address this is to extend the input domain and apply zero padding on the extended region. Specifically, one would replace the input noise with $\hat{B}^n$, an extension of a sample path of $B^n$ to the interval $[0, 2T]$, where $\hat{B}^n(t) = 0$ for $t \in (T, 2T]$. Although this method is widely used, it is not suitable in our setting. This is because in general, $\hat{B}^n$ is not continuous at $t = T$ unless $\hat{B}^n(T) = 0$, and hence does not belong to the Sobolev space $H^1([0, 2T], \mathbb{R}^d)$.

To resolve this, we employ mirror padding, wherein the input is symmetrically extended by reflecting the sample path about the midpoint $t = T$. The domain is extended from $[0, T]$ to $[0, 2T]$, and the function values on $(T, 2T]$ are defined by the mirror image of the original path. This construction yields a continuous and periodic function on $[0, 2T]$, thereby satisfying the conditions required by Theorem \ref{uatpfno}.
We recall that $\mathbb{T}=[0,2T]/\sim$ is the one-dimensional torus, $\mathcal{A}(\mathbb{T},\mathbb{R}^{d_{a}})$ is a Banach space of $\mathbb{R}^{d_{a}}$-valued continuous functions on $\mathbb{T}$,and $\mathcal{I}_{N}:C(\mathbb{T},\mathbb{R}^{d_{v}})\to C_N(\mathbb{T},\mathbb{R}^{d_{v}})$ is the pseudo-spectral Fourier projection operator.

\begin{figure}[htp!]
	\centering
	\resizebox{\columnwidth}{!}{
		\includegraphics[height=8.0cm]{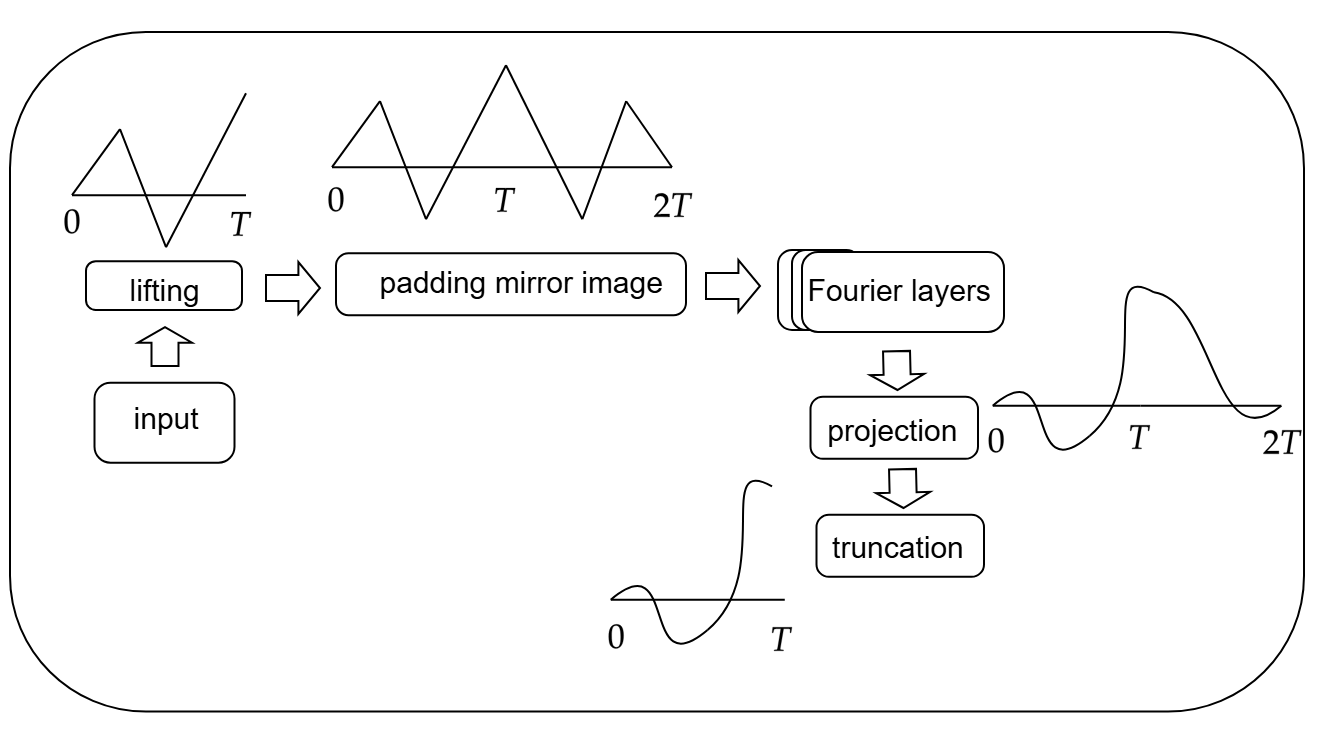}
	}
	\caption{Mirror-Padded FNO (MFNO) architecture.}\label{MFNO}
\end{figure}

\begin{definition}[1D mirror-padded FNO]
	Let $s,s'\ge 0$ and $d_{a},d_{u}\in\mathbb{N}$. Define the mirror-padding layer $\mathcal{M}:\mathcal{A}([0,T],\mathbb{R}^{d_{a}})\to \mathcal{A}(\mathbb{T},\mathbb{R}^{d_{a}})$ by 
	\[\mathcal{M}(a)(t)=\begin{cases}
	a(t) &0\le t\le T\\
	a(2T-t) &T<t\le2T
	\end{cases}\;\textnormal{ for } a\in \mathcal{A}([0,T],\mathbb{R}^{d_{a}})\] 
	and the truncating layer $\mathcal{T}:C_N(\mathbb{T}^{d},\mathbb{R}^{d_{u}})\to L^2([0,T],\mathbb{R}^{d_{u}})$ by $\mathcal{T}(u)=u|_{[0,T]}$ for $u\in C_N(\mathbb{T}^{d},\mathbb{R}^{d_{u}})$. A mirror-padded FNO (MFNO) with order $N$ is a mapping
	\[\mathcal{N}:\mathcal{A}([0,T],\mathbb{R}^{d_{a}})\to L^2([0,T],\mathbb{R}^{d_u})\]
	of the form
	\[\mathcal{N}=\mathcal{T}\circ\mathcal{Q}\circ \mathcal{I}_{N}\circ \mathcal{L}_{L}\circ \mathcal{I}_{N}\circ\cdots \circ \mathcal{L}_{1}\circ\mathcal{I}_{N}\circ\mathcal{R}\circ \mathcal{M},\]
	where $\mathcal{R}:\mathcal{A}(\mathbb{T},\mathbb{R}^{d_{a}})\to C(\mathbb{T},\mathbb{R}^{d_v})$, $\mathcal{Q}:C_N(\mathbb{T},\mathbb{R}^{d_{v}})\to C_N(\mathbb{T},\mathbb{R}^{d_{u}})$, and $\mathcal{L}_{1},\cdots,\mathcal{L}_L:C_N(\mathbb{T},\mathbb{R}^{d_{v}})\to C(\mathbb{T},\mathbb{R}^{d_{v}})$ are of the form in \eqref{lift}, \eqref{project}, and \eqref{flayer}, respectively.  
\end{definition}

The following theorem is the MFNO version of the universal approximation theorems.
\begin{theorem}[Universal Approximation for MFNOs]
	\label{uatMFNO}
	Let $s\geq \frac{1}{2}$ and $d_{a},d_{u}\in\mathbb{N}$. Suppose that \[\mathcal{G}:H^{s}([0,T],\mathbb{R}^{d_{a}})\to L^2([0,T],\mathbb{R}^{d_{u}})\] is a continuous operator and $K\subset H^{s}([0,T],\mathbb{R}^{d_{a}})$ is a compact subset. Then, for any $\epsilon>0$, there exist $N\in\mathbb{N}$ and an MFNO $\mathcal{N}:H^{s}([0,T],\mathbb{R}^{d_{a}})\to C_N(\mathbb{T},\mathbb{R}^{d_u})$ with order $N$ such that
	\[\sup_{a\in K} \|\mathcal{G}(a)-\mathcal{N}(a)\|_{L^2}\le \epsilon.\]
\end{theorem}

This result can be proven straightforwardly. For a given operator $\mathcal{G}$, we define $\tilde{\mathcal{G}}:H^{s}(\mathbb{T},\mathbb{R}^{d_{a}})\to L^2(\mathbb{T},\mathbb{R}^{d_{u}})$ by 
\[\tilde{\mathcal{G}}(f)(t)=\begin{cases}
\mathcal{G}(f|_{[0,T]})(t)&t\in[0,T]\,,\\
\mathcal{G}(f|_{[0,T]})(2T-t)&t\in(T,2T]\,.
\end{cases}\]
By Theorem~\ref{uatpfno}, there exists a $\Psi$-FNO $\mathcal{N}$ such that $\sup_{a\in K} \|\tilde{\mathcal{G}}(a)-\mathcal{N}(a)\|_{L^2}\le \epsilon$. As the MFNO is identical to the $\Psi$-FNO except for the initial mirror-padding and final truncation layers, and since $\mathcal{G}$ coincides with $\tilde{\mathcal{G}}$ when restricted to the domain $[0, T]$, the desired conclusion follows.

\section{Convergence analysis}
\label{sec:conv}

This section establishes the theoretical foundation for the convergence of our MFNO architecture. We prove its approximation capabilities for two important classes of non-Markovian processes: path-dependent SDEs and fBM.

\subsection{Path-dependent SDEs}

This subsection details the properties of path-dependent SDEs and presents the proof of our approximation theorem for MFNOs. For more details on path-dependent SDEs, we refer the reader to \cite{CF:10}. We begin with the notions of non-anticipative functionals and their derivatives. Let $D([0,T],\mathbb{R}^{m})$ be the space of $m$-dimensional c\`adl\`ag paths on $[0,T]$, equipped with the supremum norm. For $t\in[0,T]$ and $\gamma\in D([0,T],\mathbb{R}^{d})$, we denote by $\gamma(t)$ the value of $\gamma$ at time $t$ and by $\gamma_{t}$ the stopped path of $\gamma$ at $t$. Let $(e_{i})_{i=1,...,m}$ be the standard basis of $\mathbb{R}^{m}$. The indicator function is denoted by $\mathbf{1}$.
\begin{definition}
	A non-anticipative functional on $[0,T]\times D([0,T],\mathbb{R}^{m})$ is a map $$f:[0,T]\times D([0,T],\mathbb{R}^{m})\to \mathbb{R}$$ such that
	$f(t,\gamma)=f(t,\gamma_{t})$
	for all $t\in[0,T]$ and $\gamma\in D([0,T],\mathbb{R}^{m})$. It is said to be horizontally differentiable (or vertically differentiable) if for all $t\in[0,T)$ and $\gamma\in D([0,T],\mathbb{R}^{m})$, the limit
	\[\partial_{t}f(t,\gamma)=\lim_{h\to 0+}\frac{f(t+h,\gamma_{t})-f(t,\gamma)}{h}\]
	exists (or the limit
	\[\partial_{i}f(t,\gamma)=\lim_{h\to 0+}\frac{f(t+h,\gamma+he_{i}\mathbf{1}_{[t,T]})-f(t,\gamma)}{h}\]
	exists for all $i=1,\cdots,m$, respectively). We denote the vertical derivatives collectively as $\nabla f=(\partial_{1}f,\cdots,\partial_{m}f)$. A non-anticipative functional $f:[0,T]\times D([0,T],\mathbb{R}^{m})\to\mathbb{R}$ is of class $\mathcal{C}^{1,1}$ if it has continuous horizontal and vertical derivatives.
\end{definition}

We restrict the domain of non-anticipative functionals to the space of continuous functions.
If two non-anticipative functionals on $[0,T]\times D([0,T],\mathbb{R}^{m})$ are of class $\mathcal{C}^{1,1}$ and agree on all continuous paths, then their vertical derivatives also agree, as stated in \cite[Theorem 5.4.1]{BCC:16}.

\begin{definition}
	A non-anticipative functional on $[0,T]\times C([0,T],\mathbb{R}^{m})$ is a map $$f:[0,T]\times C([0,T],\mathbb{R}^{d})\to \mathbb{R}$$ such that
	$f(t,\gamma)=f(t,\gamma_{t})$
	for all $t\in[0,T]$ and $\gamma\in C([0,T],\mathbb{R}^{m})$.
	\begin{enumerate}
		\item A non-anticipative functional $f$ on $[0,T]\times C([0,T],\mathbb{R}^{d})$ is said to be of class $C^{1,1}$ if there exists a non-anticipative functional $\tilde{f}$ on 
		$[0,T]\times D([0,T],\mathbb{R}^{m})$ of class $\mathcal{C}^{1,1}$ such that $f(t,\gamma)=\tilde{f}(t,\gamma)$ for all $t\in[0,T]$ and $\gamma\in C([0,T],\mathbb{R}^{m})$.
		\item 
		For a non-anticipative functional $f$ on $[0,T]\times C([0,T],\mathbb{R}^{d})$ of class $C^{1,1}$, the horizontal and vertical derivatives of $f$ are defined as $\partial_{t}f(t,\gamma):=\partial_{t}\tilde{f}(t,\gamma)$
		and 	$\nabla f(t,\gamma):=\nabla \tilde{f}(t,\gamma)$, respectively, 
		for $t\in[0,T)$ and $\gamma\in C([0,T],\mathbb{R}^{m})$.
	\end{enumerate}
\end{definition}

We now describe path-dependent SDEs. Let $(\Omega,\mathcal{F},(\mathcal{F}_{t})_{t\in[0,T]},P)$ be a filtered probability space with an $\ell$-dimensional Brownian motion $B$. Consider the SDE of the form
\begin{equation}
\label{pdSDE}
\begin{aligned}
&  dX(t) = b(t,X)\,dt + \sigma(t,X)\,dB(t) \quad t\in[0,T],\\
& X(0)=\xi
\end{aligned}
\end{equation}
where $b:[0,T]\times C([0,T],\mathbb{R}^{m})\to \mathbb{R}^m$ and $\sigma:[0,T]\times C([0,T],\mathbb{R}^{m})\to \mathbb{R}^{m\times \ell}$ are non-anticipative functionals, and $\xi$ is an $\mathbb{R}^{m}$-valued $\mathcal{F}_{0}$-measurable random variable. This SDE is known to have a unique solution under the conditions \textbf{(R1)}-\textbf{(R3)} stated below.

We apply the Wong--Zakai approximation to demonstrate that the solution to a path-dependent SDE can be learned by an MFNO. We express a solution $X$ to \eqref{pdSDE} using the Stratonovich integral as
\begin{equation}
\label{wz pdsde}
\begin{aligned}
&dX(t)=k(t,X)\,dt+\sigma(t,X)\circ dB(t),\ \ t\in[0,T],  \\
& X(0)=\xi,
\end{aligned}
\end{equation}
where $\rho:=(\nabla\sigma^\top)\sigma$ and $k:=b-\frac{1}{2}\rho$. Notably, if $b$ and $\sigma$ satisfy conditions \textbf{(R1)}–\textbf{(R3)}, then $k$ and $\sigma$ also satisfy these conditions, with $k$ replacing $b$.

\begin{definition}
	Consider a sequence $(\pi_{n})_{n\in\mathbb{N}}$ of uniform partitions of $[0,T]$ with $|\pi_{n}|\to 0$ as $n\to\infty$. Let $B^{n}$ be the non-adapted piecewise linear interpolation of $B$ with respect to $\pi_{n}$. The Wong--Zakai approximation for the solution $X$ to \eqref{wz pdsde} is defined as the sequence of solutions $(\tilde{X}^n)_{n\in\mathbb{N}}$ to 
	\begin{equation}
	\label{WZap} 
	\begin{aligned} 
	& d\tilde{X}^n(t)=k(t,\tilde{X}^n)\,dt+\sigma(t,\tilde{X}^n)\,dB^{n}(t),\quad t\in[0,T]\,,\\ 
	&\tilde{X}^n(0)=\xi\,.
	\end{aligned}
	\end{equation} 
\end{definition}

We impose the following regularity conditions on $b$ and $\sigma$ for a fixed terminal time $T > 0$.

\begin{enumerate}[label=\textbf{(R\arabic*)}]
	\item The non-anticipative functional $\sigma$ on $[0,T)\times C([0,T],\mathbb{R}^{m})$ is of class $\mathcal{C}^{1,1}$, and there exists a positive constant $C$ such that
	\[|\sigma(t,\gamma)| + |\nabla\sigma_{k,l}(t,\gamma)|\le C\]
	for all $t\in [0,T)$, $\gamma\in C([0,T],\mathbb{R}^{m})$, $k\in\{1,\cdots,m\}$, and $l\in\{1,\cdots,\ell\}$.
	\item There exist positive constants $C$ and $\eta$ such that
	\begin{align*}
	|b(t,\gamma)|&\le C(1+\|\gamma_{t}\|_{L^{\infty}}),\\
	|\partial_{t}\sigma(t,\gamma)|&\le C(1+\|\gamma_{t}\|^{\eta}_{L^\infty})
	\end{align*}
	for all $t\in [0,T)$ and $\gamma\in C([0,T], \mathbb{R}^{m})$.
	\item There exists a positive constant $\lambda$ such that
	\begin{align*}
	|b(t,\gamma^{1})-b(t,\gamma^{2})|&\le \lambda(\|\gamma^{1}_{t}-\gamma^{2}_{t}\|_{L^{\infty}}),\\
	|\sigma(t,\gamma^{1})-\sigma(s,\gamma^{2})|&\le \lambda(|t-s|^{\frac{1}{2}}+\|\gamma^{1}_{t}-\gamma^{2}_{s}\|_{L^{\infty}}),\\
	|\nabla\sigma_{k,l}(t,\gamma^{1})-\nabla\sigma_{k,l}(s,\gamma^{2})|&\le \lambda(|t-s|^{\frac{1}{2}}+\|\gamma^{1}_{t}-\gamma^{2}_{s}\|_{L^{\infty}}).
	\end{align*}
	for all $s,t\in [0,T)$, $\gamma^{1},\gamma^{2}\in C([0,T],\mathbb{R}^{d})$, $k\in\{1,\cdots,m\}$, and $l\in\{1,\cdots,\ell\}$.
\end{enumerate}

The following theorem, a direct result of \cite[Theorem 3.1]{XG:23}, provides an estimate of the difference between the original solution $X$ and its Wong--Zakai approximation $\tilde{X}^n$.

\begin{theorem}
	\label{thm pdwz}
	Suppose that \textbf{(R1)}-\textbf{(R3)} and $\mathbb{E}|\xi|^{q}<\infty$ for all $q\ge 2$ hold. Let $X$ and $\tilde{X}^{n}$ be the solutions to \eqref{wz pdsde} and \eqref{WZap}, respectively. Then, for every $p>2$, there exists a constant $C_{p}>0$ such that 
	\[(\mathbb{E}\|X-\tilde{X}^n\|^{p}_\infty)^{\frac{1}{p}} \le C_{p}|\pi_{n}|^{\frac{1}{2}-\frac{1}{p}}\]
	for all $n\in\mathbb{N}.$
\end{theorem}

We now state an approximation theorem of MFNOs for solutions to \eqref{pdSDE}. We slightly abuse notation by identifying a constant function in $H^{1}([0,T],\mathbb{R}^{m})$ with its value in $\mathbb{R}^{m}$ and vice versa.

\begin{theorem}
	\label{mainthm}
	Suppose that $b$ and $\sigma$ satisfy \textbf{(R1)-(R3)}.    
	Let $\xi$ be a bounded $\mathcal{F}_{0}$-measurable random variable and  $X^{\xi}$ be the solution to \eqref{pdSDE}. Moreover, let $B^n$ be the non-adapted piecewise linear interpolation of $B$ with respect to a uniform partition $\pi_n$ satisfying $|\pi_n|\to0$ as $n\to\infty.$     Then, for any $\epsilon,\epsilon'>0$ and $M\in \mathbb{N}$, there exist $N,M_{0}\in\mathbb{N}$, a subset $D$ of $\Omega$ and a MFNO $\mathcal{N}: H^1([0,T],\mathbb{R}^{m+\ell})\to L^{2}([0,T],\mathbb{R}^{m})$ with order $N$ such that $\mathbb{P}(D)>1-\epsilon'$ and
	\[(\mathbb{E}\left[\|X^{\xi}-\mathcal{N}(\xi,B^{n})\|^{2}_{L^{\infty}}\mid D\right])^{\frac{1}{2}}<\epsilon\]
	whenever $M_0\le n \le M_0+M$.
\end{theorem}

We prove this theorem in several steps.
For $\gamma\in H^{1}([0, T],\mathbb{R}^{m})$ and $\omega\in H^{1}([0,T],\mathbb{R}^{d})$, we consider the ODE
\begin{equation}
\label{WZap1}
\begin{aligned}
&d\tilde{X}(t)=b(t,\tilde{X})\,dt+\sigma(t,\tilde{X})\,d\omega(t),\quad t\in[0,T]\,,\\
&\tilde{X}(0)=\gamma(0)\,.
\end{aligned}
\end{equation}
We denote its solution as $\tilde{X}^{\gamma,\omega}$.

\begin{lemma}  Suppose that \textbf{(R1)}-\textbf{(R3)} hold. Then, there is a unique solution $\tilde{X}^{\gamma,\omega}$ to \eqref{WZap1} in $H^{1}([0,T],\mathbb{R}^{m})$.
\end{lemma}
\begin{proof} We use the Banach fixed-point theorem. For $X\in H^{1}([0,\delta],\mathbb{R}^{m})$ and $\delta>0$, define $\Phi(X)$ as
	\[\Phi(X)(t) =
	\gamma(0)+ \int^{t}_{0} b(t,X)\,dt +\int^{t}_{0} \sigma(t,X)\,d\omega(t),\quad t\in [0,\delta].
	\]
	Then, for any $X,Y\in H^{1}([-T,\delta],\mathbb{R}^{m})$, we have
	\begin{align*}
	\|\Phi(X)'-\Phi(Y)'\|^{2}_{L^{2}} &= \int^{\delta}_{0}|b(t,X)-b(t,Y) +(\sigma(t,X)-\sigma(t,Y))\omega'(t)|^{2}\,dt\\
	&\le \int^{\delta}_{0}\big(\lambda\|X-Y\|_{L^{\infty}}+ \lambda\|X-Y\|_{L^{\infty}}|\omega'(t)|\big)^{2}\,dt\\
	&\le \lambda^{2}\|X-Y\|^{2}_{L^{\infty}}\int^{\delta}_{0}\big(1+\|\omega'\|_{L^{\infty}}\big)^{2}\,dt\\ 
	&\le \delta\lambda^{2}(\tfrac{1}{T}+T)\|X-Y\|^{2}_{H^{1}}(1+\sqrt{\tfrac{1}{T}+T}\,\|\omega'\|_{H^{1}})^{2}.
	\end{align*}
	Furthermore, since $\Phi(X)(0)-\Phi(Y)(0)=0$, we have 
	\[\|\Phi(X)-\Phi(Y)\|^{2}_{L^{2}}\le \frac{\delta^{2}}{2}\|\Phi(X)'-\Phi(Y)'\|^{2}_{L^{2}}.\]
	From the above inequalities, we obtain
	\begin{align*} 
	\|\Phi(X)-\Phi(Y)\|^{2}_{H^{1}}
	&=\|\Phi(X)-\Phi(Y)\|^{2}_{L^{2}}+\|\Phi(X)'-\Phi(Y)'\|^{2}_{L^{2}}\\
	&\le \delta(1+\tfrac{\delta^{2}}{2})\lambda^{2}(\tfrac{1}{T}+T)(1+\sqrt{\tfrac{1}{T}+T}\|\omega'\|_{H^{1}})^{2}\|X-Y\|^{2}_{H^{1}}\,.  
	\end{align*}
	Thus, for $$0<\delta<\min\Big(\tfrac{1}{2}, \frac{1}{2\lambda^{2}(\tfrac{1}{T}+T)(1+\sqrt{\tfrac{1}{T}+T}\|\omega'\|_{H^{1}})^{2}}\Big)\,,$$ the map $\Phi$ is a contraction on $H^{1}([-T,\delta],\mathbb{R}^{m})$. Therefore, by the Banach fixed-point theorem and the standard pasting argument, we obtain the desired result. 
	\qed
\end{proof}

\begin{lemma}
	Define an operator  $F:H^{1}([0,T],\mathbb{R}^{m+d})\to H^{1}([0,T],\mathbb{R}^{m})$ by
	$F(\gamma,\omega)=\tilde{X}^{\gamma,\omega}$. Then, the map $F$ is continuous.
\end{lemma}

\begin{proof} Let $(\gamma^{1},\omega^{1}),(\gamma^{2},\omega^{2})\in H^{1}([0,T],\mathbb{R}^{d+m})$ and let $X^{1}=F(\gamma^{1},\omega^{1})$ and $X^{2}=F(\gamma^{2},\omega^{2})$. Then, for $t\in[0,T]$,
	\begin{align*}
	(X^{1})'(t) - (X^{2})'(t)
	&= b(t, X^{1}) - b(t, X^{2}) \\
	&\quad + \left( \sigma(t, X^{1}) - \sigma(t, X^{2}) \right)(\omega^{1})'(t) \\
	&\quad + \sigma(t, X^{2}) \left( (\omega^{1})'(t) - (\omega^{2})'(t) \right).
	\end{align*}
	Thus, 
	\[|(X^{1})'(t)-(X^{2})'(t)|\le \lambda \|X^{1}_{t}-X^{2}_{t}\|_{L^{\infty}}(1+|(\omega^{1})'(t)|) + C|(\omega^{1})'(t)-(\omega^{2})'(t)|.\]
	For each $N\in\mathbb{N}$,
	we consider the partition $\{t_k:=\frac{kT}{N}\,|\,k=0,1,\cdots,N\}$ of $[0,T]$.
	We estimate the $H^1([t_k,t_{k+1}],\mathbb{R}^m)$-norm of $X^1-X^2$ for each $k=0,1,\cdots,N.$
	Observe that
	\begin{equation} \label{3}
	\begin{aligned}
	&\quad\|(X^1)' - (X^2)'\|^2_{L^2([t_k, t_{k+1}])}\\
	&= \int_{t_k}^{t_{k+1}} |(X^1)'(t) - (X^2)'(t)|^2 \, dt \\
	&\le 2\lambda^2 \|X^1 - X^2\|^2_{L^\infty([0, t_{k+1}])}  \int_{t_k}^{t_{k+1}} \left(1 + |(\omega^1)'(t)| \right)^2 dt \\
	&\quad + 2C^2 \int_{t_k}^{t_{k+1}} |(\omega^1)'(t) - (\omega^2)'(t)|^2 dt \\
	&\le \frac{2T}{N} \lambda^2 \left( \tfrac{1}{T} + T \right)
	\left( 1 + \sqrt{\tfrac{1}{T} + T} \|\omega^1\|_{H^1} \right) \|X^1 - X^2\|^2_{H^1([0, t_{k+1}])}\\
	&\quad 
	+ 2C^2 \|\omega^1 - \omega^2\|^2_{H^1} \\
	&= \frac{A}{N} \|X^1 - X^2\|^2_{H^1([0, t_{k+1}])}
	+ B \|\omega^1 - \omega^2\|^2_{H^1},
	\end{aligned}
	\end{equation}
	where $A:=2T\lambda^{2}(\tfrac{1}{T}+T)(1+\sqrt{\tfrac{1}{T}+T}\|\omega^{1}\|_{H^{1}})$ and $B:=2C^{2}$.
	Furthermore, using
	\begin{align*}
	|X^{1}(t)-X^{2}(t)|
	&\le|X^{1}(t)-X^{2}(t)-(X^{1}(t_k)-X^{2}(t_k))|+|X^{1}(t_k)-X^{2}(t_k)|\\
	&\le\int_{t_k}^{t_{k+1}}|(X^1)'(t)-(X^2)'(t)|\,dt+|X^{1}(t_k)-X^{2}(t_k)|\,,
	\end{align*}
	we have
	\begin{align*}
	&\quad\,\,\|X^1-X^2\|^2_{L^2([t_k,t_{k+1}],\mathbb{R}^m)}=\int^{t_{k+1}}_{t_k}|X^{1}(t)-X^{2}(t)|^{2}dt\\
	&\le \big(\frac{T}{N}\big)^{2}\int^{t_{k+1}}_{t_{k}}|(X^{1})'(t)-(X^{2})'(t)|^{2}dt+\frac{2T}{N}|X^{1}(t_{k})-X^{2}(t_{k})|^{2}\\
	&\le \frac{T^{2}A}{N^{3}}\|X^{1}-X^{2}\|^{2}_{H^{1}([0,t_{k+1}])} + \frac{BT^{2}}{N^{2}}\|\omega^{1}-\omega^{2}\|^{2}_{H^{1}}+\frac{2T}{N}|X^{1}(t_{k})-X^{2}(t_{k})|^{2}
	\end{align*}
	Along with inequality \eqref{3}, we obtain
	\begin{align*}
	&\quad\,\,\|X^{1}-X^{2}\|^{2}_{H^{1}([t_{k},t_{k+1}])}\\
	&=\|X^1-X^2\|^2_{L^2([t_k,t_{k+1}],\mathbb{R}^m)}+ \|(X^1)'-(X^2)'\|^2_{L^2([t_k,t_{k+1}],\mathbb{R}^m)}\\
	&\le (\frac{A}{N}+\frac{T^{2}A}{N^{3}})\|X^{1}-X^{2}\|^{2}_{H^{1}([0,t_{k+1}])} + (B+\frac{BT^{2}}{N^{2}})\|\omega^{1}-\omega^{2}\|^{2}_{H^{1}}\\
	&\quad +\frac{2T}{N}|X^{1}(t_{k})-X^{2}(t_{k})|^{2}\\  
	&\le (\frac{A}{N}+\frac{T^{2}A}{N^{3}})\|X^{1}-X^{2}\|^{2}_{H^{1}([0,t_{k}])}+(\frac{A}{N}+\frac{T^{2}A}{N^{3}})\|X^{1}-X^{2}\|^{2}_{H^{1}([t_{k},t_{k+1}])} \\
	&\quad+ (B+\frac{BT^{2}}{N^{2}})\|\omega^{1}-\omega^{2}\|^{2}_{H^{1}}+\frac{2T}{N}|X^{1}(t_{k})-X^{2}(t_{k})|^{2}.
	\end{align*}
	It follows that
	\begin{align*}
	&\quad\;\|X^{1}-X^{2}\|^{2}_{H^{1}([t_{k},t_{k+1}])}\\
	&\le \frac{A/N+T^{2}A/N^{3}}{1-(\frac{T^{2}A}{N^{3}}+\frac{A}{N})}\|X^{1}-X^{2}\|^{2}_{H^{1}([0,t_{k}])}+\frac{B+BT^{2}/N^{2}}{1-(\frac{T^{2}A}{N^{3}}+\frac{A}{N})}\|\omega^{1}-\omega^{2}\|^{2}_{H^{1}}\\
	&\quad +\frac{2T}{N\big(1-(\frac{T^{2}A}{N^{3}}+\frac{A}{N})\big)}|X^{1}(t_{k})-X^{2}(t_{k})|^{2} \\
	&\le \frac{1}{2}\|X^{1}-X^{2}\|^{2}_{H^{1}([0,t_{k}])}+4B\|\omega^{1}-\omega^{2}\|^{2}_{H^{1}}+\frac{4T}{N}|X^{1}(t_{k})-X^{2}(t_{k})|^{2} 
	\end{align*}
	by choosing $N\ge \max(8T^{2}A,8A,T)$.

	For $k=0$, choosing $N\ge \max(8T^{2}A,8A,T,4(1+T^{2}))$, we have  
	\begin{equation}
	\begin{aligned}
	\|X^{1}-X^{2}\|_{H^{1}([0,t_{1}])}&\le 4B\|\omega^{1}-\omega^{2}\|^{2}_{H^{1}}+\frac{4T}{N}|X^{1}(0)-X^{2}(0)|^{2}\\
	&\le 4B\|\omega^{1}-\omega^{2}\|^{2}_{H^{1}}+\frac{4T}{N}(\frac{1}{T}+T)\|\gamma^{1}-\gamma^{2}\|^{2}_{H^{1}}\\ 
	&\le 4B\|\omega^{1}-\omega^{2}\|^{2}_{H^{1}}+\|\gamma^{1}-\gamma^{2}\|^{2}_{H^{1}}
	\end{aligned}
	\end{equation} 
	where we used the fact that
	$|X^{1}(0)-X^{2}(0)|=|\gamma^{1}(0)-\gamma^{2}(0)|\le (\frac{1}{T}+T)\|\gamma^{1}-\gamma^{2}\|^{2}_{H^{1}}$.
	For $k\in\{1,\cdots,N-1\}$, applying 
	\begin{align*}
	|X^{1}(t_{k})-X^{2}(t_{k})|^{2}&\le (\tfrac{N}{T}+\tfrac{T}{N})\|X^{1}-X^{2}\|^{2}_{H^{1}([\frac{T(k-1)}{N},t_{k}])}\\
	&\le (\tfrac{N}{T}+\tfrac{T}{N})\|X^{1}-X^{2}\|^{2}_{H^{1}([0,t_{k}])}
	\end{align*}
	yields 
	\begin{align*}
	&\|X^{1}-X^{2}\|^{2}_{H^{1}([t_{k},t_{k+1}])}\\
	&\le4B\|\omega^{1}-\omega^{2}\|^{2}_{H^{1}}+(\frac{1}{2}+\frac{4T}{N}(\frac{N}{T}+\frac{T}{N}))\|X^{1}-X^{2}\|^{2}_{H^{1}([0,t_{k}])}\\
	&\le4B\|\omega^{1}-\omega^{2}\|^{2}_{H^{1}}+5\|X^{1}-X^{2}\|^{2}_{H^{1}([0,t_{k}])}\\
	&= 4B\|\omega^{1}-\omega^{2}\|^{2}_{H^{1}}+ 5\sum^{k-1}_{i=0}\|X^{1}-X^{2}\|^{2}_{H^{1}([t_{i},t_{i+1}])}.
	\end{align*}
	Now, consider a sequence $\{a_{n}\}_{n\in\mathbb{N}_{0}}$ defined by
	\begin{align*}
	a_{0}&=4B\|\omega^{1}-\omega^{2}\|^{2}_{H^{1}}+\|\gamma^{1}-\gamma^{2}\|^{2}_{H^{1}},\\
	a_{k}&=4B\|\omega^{1}-\omega^{2}\|^{2}_{H^{1}}+ 5\sum^{k-1}_{i=0} a_{i}\ \ \ (k\ge1).
	\end{align*}
	Clearly, $\|X^{1}-X^{2}\|^{2}_{H^{1}([t_{k},t_{k+1}])}\le a_{k}$ for all $k\in\{0,\cdots,N-1\}$. The linear recurrence relation for $\{a_{n}\}_{n\in\mathbb{N}_{0}}$ yields a closed-form solution, from which we obtain
	\[\sum^{N-1}_{i=0} a_{k}=\frac{4B(6^{N}-1)}{5}\|\omega^{1}-\omega^{2}\|^{2}_{H^{1}}+6^{N-1}\|\gamma^{1}-\gamma^{2}\|^{2}_{H^{1}}.\]
	Thus, for any $\epsilon>0$, if $\|(\gamma^{1}-\gamma^{2}, \omega^{1}-\omega^{2})\|_{H^{1}}< \sqrt{\min(\frac{\epsilon}{2(6^{N}-1)}, \frac{5\epsilon}{8B(6^{N}-1)})} $, we obtain
	\[\|X^{1}-X^{2}\|^{2}_{H^{1}([0,T])}=\sum^{N-1}_{k=0}\|X^{1}-X^{2}\|^{2}_{H^{1}([t_{k},t_{k+1}])}\le \sum^{N-1}_{k=0} a_{k}<\epsilon.\]
	Therefore, the operator $F:H^{1}([0,T],\mathbb{R}^{d+m})\to H^{1}([0,T],\mathbb{R}^{m})$ is continuous.
	\qed
\end{proof}

\begin{proof}[Proof of Theorem~\ref{mainthm}]
	By Theorem~\ref{thm pdwz}, for any $p>2$, there exists a constant $C_{p}>0$ such that
	\[(\mathbb{E}\|X^{\xi}-F(\xi,B^{n})\|^{p}_{L^\infty})^{\frac{1}{p}} \le C_{p}|\pi_{n}|^{\frac{1}{2}-\frac{1}{p}}\]
	for all $n\in\mathbb{N}$. Since $|\pi_{n}|\to0$ as $n\to \infty$, there exists an $M_{0}\in\mathbb{N}$ such that 
	\[(\mathbb{E}\|X^{\xi}-F(\xi,B^{n})\|^{p}_{L^\infty})^{\frac{1}{p}} <\frac{\epsilon(1-\epsilon')^{\frac{1}{p}}}{2}.\]
	for all $n\ge M_{0}$.
	
	We denote by $\mathcal{P}\ell(\pi_{n},\mathbb{R}^{\ell})$, the finite-dimensional subspace of $H^{1}([0,T],\mathbb{R}^{\ell})$ consisting of functions that are piecewise linear with respect to the partition $\pi_{n}$. Choose $R>0$ such that
	$\mathbb{P}(|\xi|\le R)=1$. Let  
	\[
	\begin{aligned}
	K := \Big\{ (x, \omega) \in H^{1}([0,T], \mathbb{R}^{m+\ell}) \,\Big|\,
	& \|x\|_{H^{1}} \le R,\ \|\omega\|_{H^{1}} \le R_{M_{0}+M, \epsilon'}, \\
	& \omega \in \mathcal{P}\ell(\pi_{M_{0}+M}, \mathbb{R}^{\ell}) \Big\}
	\end{aligned}
	\]
	and let
	$$D:=\{\omega\in \Omega\,|\,(\xi(\omega),B^{n}(\omega))\in K 
	\textnormal{ for }  M_{0}\le n\le M_{0}+M\},$$ where $R_{M_{0}+M,\epsilon'}$ is the constant given in Lemma~\ref{lemmaPB}. Then, $K$ is a compact subset of $H^{1}([0,T],\mathbb{R}^{m+\ell})$, and by Lemma~\ref{lemmaPB}, we have $\mathbb{P}(D)>1-\epsilon'$,
	which implies
	\[(\mathbb{E}\left[\|X^{\xi}-F(\xi,B^{n})\|^{p}_{L^\infty}\mid D\right])^{\frac{1}{p}} <\frac{\epsilon(1-\epsilon')^{\frac{1}{p}}}{2(1-\epsilon')^{\frac{1}{p}}}=\frac{\epsilon}{2}\]
	whenever $M_{0}\le n\le M_{0}+M$. Thus,
	\[(\mathbb{E}\left[\|X^{\xi}-F(\xi,B^{n})\|^{2}_{L^\infty}\mid D\right])^{\frac{1}{2}}\le (\mathbb{E}\left[\|X^{\xi}-F(\xi,B^{n})\|^{p}_{L^\infty}\mid D\right])^{\frac{1}{p}}<\frac{\epsilon}{2}\]
	whenever $M_{0}\le n\le M_{0}+M$. 
	
	By the universal approximation theorem for MFNOs (Theorem~\ref{uatMFNO}), there exists an MFNO $\mathcal{N}:H^1([0,T],\mathbb{R}^{m+\ell})\to L^{2}([0,T],\mathbb{R}^{m})$ with order $N$ such that
	\[\sup_{(x,\omega)\in K}\|F(x,\omega)-\mathcal{N}(x,\omega)\|_{H^{1}}<\frac{\epsilon}{2\sqrt{\tfrac{1}{T}+T}},\]
	so that 
	\[\|F(\xi,B^{n})-\mathcal{N}(\xi,B^{n})\|_{L^{\infty}}<\frac{\epsilon}{2}\]
	on the set $D$. Thus, for $M_{0}\le n\le M_{0}+M$,
	\begin{align*}
	&\left(\mathbb{E}\left[\|X^{\xi}-\mathcal{N}(\xi,B^{n})\|^{2}_{L^{\infty}}\mid D\right]\right)^{\frac{1}{2}}\\
	&\le \left(\mathbb{E}\left[\left(\|X^{\xi}-F(\xi,B^{n})\|_{L^{\infty}}+\|F(\xi,B^{n})-\mathcal{N}(\xi,B^{n})\|_{L^{\infty}}\right)^{2}\mid D\right]\right)^{\frac{1}{2}}\\
	&\le \left(\mathbb{E}\left[\|X^{\xi}-F(\xi,B^{n})\|^{2}_{L^{\infty}}\mid D\right ]\right)^{\frac{1}{2}}+\left(\mathbb{E}\left[\|F(\xi,B^{n})-\mathcal{N}(\xi,B^{n})\|_{L^{\infty}}^{2}\mid D \right]\right)^{\frac{1}{2}}\\
	&< \frac{1}{2}\epsilon+\frac{1}{2}\epsilon\\
	&=\epsilon.
	\end{align*}
	This completes the proof. 
\end{proof}

\subsection{Fractional Brownian motion}

We now review the concept of fractional Brownian motion (fBM). Unlike standard Brownian motion, fBM allows dependent increments, rendering it well-suited for modeling non-Markovian dynamics.

\begin{definition}
	A fractional Brownian motion $B_{H}$ on $[0,T]$ with Hurst index $H\in (0,1)$ is a continuous Gaussian process such that
	\begin{enumerate}
		\item $B_{H}(0)= 0$,
		\item $\mathbb{E}[B_{H}(t)]=0$ for all $t\in[0,T]$, 
		\item $\mathbb{E}[B_{H}(t)B_{H}(s)]=\frac{1}{2}(t^{2H}+s^{2H}-|t-s|^{2H})$ for all $s,t\in [0,T]$.
	\end{enumerate}
\end{definition}

The standard Brownian motion is a special case of fBM with Hurst index $H=0.5$. An fBM has an It\^o integral representation. Let $B$ be a standard Brownian motion, and let $\Gamma$ and ${{}_2}{F}_{1}$ denote the Euler gamma function and the hypergeometric function, respectively. It is well-known that the process defined by
\[B_{H}(t):= \int^{t}_{0} K_{H}(t,s) \,dB(s)\,,\;0\le t\le T\]
is an fBM with Hurst parameter $H$, where the kernel $K_H$ is
\[K_{H}(t,s) = \frac{(t-s)^{H-\frac{1}{2}}}{\Gamma(H+\frac{1}{2})}\, {{}_2}F_{1}(H-\frac{1}{2},\frac{1}{2}-H, H+\frac{1}{2}, 1- \frac{t}{s})\,.\] 
As a preliminary,
we present the following  proposition, which corresponds to \cite[Proposition 3.1]{DU:99}.
\begin{proposition} 
	\label{fbmprop}
	A sequence of processes $(W^{n})_{n\in\mathbb{N}_{0}}$, defined by
	\begin{equation}
	\label{fbmapp}
	\begin{aligned} 
	W^{n}(t)
	&=\int^{T}_{0} K_{H}(t,s)\,dB^{n}(s)\\
	&=\sum_{t^{n}_{i}\in\pi_{n}} \frac{1}{t^{n}_{{i+1}}-t^{n}_{i}}\int^{t^{n}_{i+1}}_{t^{n}_{i}} K_{H}(t,s)\,ds\left(B(t^{n}_{i+1})-B(t^{n}_{i})\right),      
	\end{aligned}
	\end{equation}
	converges to $B_{H}$ in $L^{2}(\mathbb{P}\otimes ds)$.
\end{proposition}

We next show that the expressive capacity of MFNOs extends to fBMs, enabling the approximation of continuous operators defined on such processes. Theorem \ref{mainthm1} is one of our main results and is proved in several steps. For any $\omega\in H^{1}([0,T],\mathbb{R})$, the map $t\mapsto \int^{T}_{0} K_{H}(t,s)\,d\omega(s)$ is continuous and therefore belongs to $L^{2}([0,T],\mathbb{R})$. We consider an operator $G:H^{1}([0,T],\mathbb{R})\to L^{2}([0,T],\mathbb{R})$ defined by
\[G(\omega)(t)=\int^{T}_{0} K_{H}(t,s)\,d\omega(s).\]

\begin{lemma}
	The operator $G:H^{1}([0,T],\mathbb{R})\to L^{2}([0,T],\mathbb{R})$ is continuous.
\end{lemma}

\begin{proof}
	Let $K_H$ be the kernel with Hurst index $H\in (0,1)$. The Cauchy--Schwarz inequality yields 
	\[|G(\omega)(t)|\le \left(\int^{T}_{0} |K_{H}(t,s)|^{2} ds\right)^{1/2}\|\omega'\|_{L^{2}}.\]
	By squaring both sides and integrating over 
	$[0,T]$, we obtain \begin{equation}
	\label{fbmcontinineq}
	\|G(\omega)\|_{L^{2}}\le \left(\int^{T}_{0}\int^{T}_{0}|K_{H}(t,s)|^{2}dsdt\right)^{1/2}\|\omega\|_{H^{1}}\,.
	\end{equation}
	According to~\cite[Theorem 3.2]{DU:99},
	there exists a positive constant $c_{H}$ such that for all $t,s\ge0$, we have
	\[|K_{H}(t,s)|\le c_{H} s^{-|H-1/2|}(t-s)^{-(1/2-H)_{+}}\mathbf{1}_{[0,t]}(s)\,,\]
	where $x_{+}=\max(x,0)$.

	We consider two cases separately.
	
	\begin{enumerate}[ leftmargin=*, label=(\roman*)]
		\item Suppose $H\ge\frac{1}{2}.$
		Then, the kernel satisfies
		\[|K_{H}(t,s)|\le c_{H}s^{\frac{1}{2}-H}\mathbf{1}_{[0,t]}(s)\]
		for all $t,s\ge0$. Hence,
		\begin{align*}\int^{T}_{0}\int^{T}_{0}|K_{H}(t,s)|^{2}\,ds\,dt&\le\int^{T}_{0}\left(\int^{t}_{0}c^{2}_{H}s^{1-2H}\,ds\right)dt\\
		&\le\int^{T}_{0} \frac{c^{2}_{H}}{2-2H}t^{2-2H} \,dt \\
		&= \frac{c^{2}_{H}T^{3-2H}}{(2-2H)(3-2H)}<\infty.
		\end{align*}
		\item Suppose $H < \frac{1}{2}$. 
		Similarly, we have
		\[|K_{H}(t,s)|\le c_{H}s^{H-\frac{1}{2}}(t-s)^{H-\frac{1}{2}}\mathbf{1}_{[0,t]}(s)\]
		for all $t,s\ge0$. Hence,
		\[\int^{T}_{0}\int^{T}_{0}|K_{H}(t,s)|^{2}\,ds\,dt\le\int^{T}_{0}\left(\int^{t}_{0}c^{2}_{H}s^{2H-1}(t-s)^{2H-1}\,ds\right)dt.\]
		Substituting $s=tu$, we find
		\begin{align*}
		\int^{t}_{0}s^{2H-1}(t-s)^{2H-1}\,ds&=\int^{1}_{0}(tu)^{2H-1}(t-tu)^{2H-1}t\, du\\
		&=t^{4H-1}\int^{t}_{0}u^{2H-1}(1-u)^{2H-1}\,du\\
		&=t^{4H-1}\frac{\Gamma(2H)^{2}}{\Gamma(4H)}.
		\end{align*}
		where  $\Gamma$ is the Gamma function. Thus, we obtain
		\begin{align*}
		\int^{T}_{0}\int^{T}_{0}|K_{H}(t,s)|^{2}dsdt&\le \int^{T}_{0} c^{2}_{H}t^{4H-1}\frac{\Gamma(2H)^{2}}{\Gamma(4H)} dt\\&=\frac{c^{2}_{H}T^{4H}\Gamma(2H)^{2}}{4H\,\Gamma(4H)}<\infty.
		\end{align*}
	\end{enumerate}
	\noindent In both cases, there exists a constant $C_{H}>0$ such that
	\[\|G(\omega)\|_{L^{2}}\le C_{H}\|\omega\|_{H^{1}}\]
	for all $\omega\in H^{1}([0,T],\mathbb{R})$
	and, therefore, $G$ is continuous.
	\qed
\end{proof}

We now state the approximation theorem for Lipschitz transformations of fBMs. For a constant $L>0,$ 
we say an operator 
$\mathcal{G}:L^{2}([0,T],\mathbb{R})\to L^{2}([0,T],\mathbb{R})$ is $L$-Lipschitz if 
$\|\mathcal{G}(a_1)-\mathcal{G}(a_2)\|_{L^2}\leq L\|a_1-a_2\|_{L^2}$
for all $a_1,a_2\in L^{2}([0,T],\mathbb{R})$. 

\begin{theorem}
	Let $\mathcal{G}:L^{2}([0,T],\mathbb{R})\to L^{2}([0,T],\mathbb{R})$ be an $L$-Lipschitz operator, and suppose $B_{H}$ is an fBM with Hurst parameter $H\in(0,1)$. Then, for any $\epsilon, \epsilon'>0$ and $M\in \mathbb{N}$, there exist $N,M_{0}\in\mathbb{N}$, a set $D$ with $\mathbb{P}(D)>1-\epsilon'$, and an MFNO $\mathcal{N}:H^{1}([0,T],\mathbb{R})\to L^{2}([0,T],\mathbb{R})$ with order $N$ such that
	\[(\mathbb{E}\left[\|\mathcal{G}(B_{H})-\mathcal{N}(B^{n})\|^{2}_{L^{2}}\mid D\right])^{\frac{1}{2}}<\epsilon\]
	whenever $M_{0}\le n\le M_{0}+M$.
\end{theorem}
\begin{proof}
	Since $\mathcal{G}$ is $L$-Lipschitz, we have
	$$\|\mathcal{G}(B_{H})-\mathcal{G}\circ G(B^{n})\|^{2}_{L^{2}}\le L^{2}\|B_{H}-G(B^{n})\|^{2}_{L^{2}},$$
	and $\mathcal{G}\circ G:H^{1}([0,T],\mathbb{R})\to L^{2}([0,T],\mathbb{R})$ is a continuous operator. By Proposition~\ref{fbmprop}, there exists an $M_{0}\in\mathbb{N}$ such that 
	\[\left(\mathbb{E}\|\mathcal{G}(B_{H})-\mathcal{G}\circ G(B^{n})\|^{2}_{L^{2}}\right)^{1/2}
	\le L\left(\mathbb{E}\|B_{H}-G(B^{n})\|^{2}_{L^{2}}\right)^{1/2}
	<\epsilon(1-\epsilon')^{1/2}/2\]
	whenever $n\ge M_{0}$.	Similar to the proof of Theorem~\ref{mainthm}, let 
	\[K=\{\omega\in H^{1}([0,T],\mathbb{R})\mid  \|\omega\|_{H^{1}}\le R_{M_{0}+M,\epsilon'}, \omega\in \mathcal{P}\ell(\pi_{M_{0}+M},\mathbb{R})\}\]
	and $D=\{\omega\in \Omega\,|\,B^{n}(\omega) \in K \textnormal{ for } M_{0}\le n\le M_{0}+M\}$, where $R_{M_{0}+M,\epsilon'}$ is the constant from Lemma~\ref{lemmaPB}. Then, $K$ is a compact subset of $H^{1}([0,T])$, and $\mathbb{P}(D)>1-\epsilon'$. 
	In addition, 
	\[(\mathbb{E}\left[\|\mathcal{G}(B_{H})-\mathcal{G}\circ G(B^{n})\|^{2}_{L^{2}}\mid D\right])^{\frac{1}{2}} <\frac{\epsilon}{2}\]
	whenever $M_{0}\le n\le M_{0}+M$. 
	
	By the universal approximation theorem for MFNOs (Theorem~\ref{uatMFNO}), there exists an MFNO $\mathcal{N}:H^{1}([0,T],\mathbb{R})\to L^{2}([0,T],\mathbb{R})$ with order $N$ such that
	\[\sup_{\omega\in K}\|\mathcal{G}\circ G(\omega)-\mathcal{N}(\omega)\|_{H^{1}}<\frac{\epsilon}{2},\]
	which implies	\[\|\mathcal{G}\circ G(B^{n})-\mathcal{N}(B^{n})\|_{L^{2}}<\frac{\epsilon}{2}\]
	on the set $D$.
	Thus, for $M_{0}\le n\le M_{0}+M$,
	\begin{align*}
	\left( \mathbb{E}\left[ \|\mathcal{G}(B_H) - \mathcal{N}(B^n)\|^2_{L^2} \mid D \right] \right)^{1/2}
	&\le \left( \mathbb{E} \left[ \left(
	\|\mathcal{G}(B_H) - \mathcal{G} \circ G(B^n)\|_{L^2}
	\right. \right. \right. \\
	&\quad \left. \left. \left. +
	\|\mathcal{G} \circ G(B^n) - \mathcal{N}(B^n)\|_{L^2}
	\right)^2 \mid D \right] \right)^{1/2} \\
	&\le \left( \mathbb{E} \left[ \|\mathcal{G}(B_H) - \mathcal{G} \circ G(B^n)\|^2_{L^2} \mid D \right] \right)^{1/2} \\
	&\quad + \left( \mathbb{E} \left[ \|\mathcal{G} \circ G(B^n) - \mathcal{N}(B^n)\|^2_{L^2} \mid D \right] \right)^{1/2} \\
	&< \tfrac{1}{2} \epsilon + \tfrac{1}{2} \epsilon \\
	&= \epsilon.
	\end{align*}
	This completes the proof.
	\qed
\end{proof}

\begin{corollary}
	\label{mainthm1}
	Let $B_{H}$ be an fBM with Hurst parameter $H\in(0,1)$. Then, for any $\epsilon,\epsilon'>0$ and $M\in \mathbb{N}$, there exist $N, M_{0}\in\mathbb{N}$, a set $D$ of $\Omega$ with $\mathbb{P}(D)>1-\epsilon'$, and an MFNO $\mathcal{N}: H^1([0,T],\mathbb{R})\to L^{2}([0,T],\mathbb{R})$ with order $N$ such that 
	\[(\mathbb{E}\left[\|B_{H}-\mathcal{N}(B^{n})\|^{2}_{L^{2}}\mid D\right])^{\frac{1}{2}}<\epsilon\]
	whenever $M_{0}\le n \le M_{0}+M$.
\end{corollary}

\section{Experiments}
\label{sec:exper}

In this section, we conduct a series of experiments to demonstrate that MFNO can effectively approximate the solutions of path-dependent SDEs and fBMs. We compare the test accuracy and inference speed of our models against several existing architectures. Finally, we analyze the resolution generalization capabilities of MFNO, ZFNO, and FNO across various tasks.

\subsection{Training algorithm}

When the underlying dynamics of the stochastic process are known, we can generate sample paths using classical simulation methods. Each generated sample path $X^{(i)}$ corresponds to a realization $B^{(i)}$ of the driving Brownian motion $B$ and an initial condition $\xi^{(i)}$, related through an operator $F$ that characterizes the system. We use these sample paths to train the MFNO, denoted $F_{\theta}$, in a supervised learning framework using regression. The model parameters are optimized by minimizing the mean squared error ($L^2$-norm loss), ensuring that the model's outputs closely match the ground-truth sample paths. The details of this training procedure are provided in Algorithm \ref{Regression}.

\begin{algorithm}
	\caption{Training with Sample Paths from Known Dynamics}\label{Regression}
	\begin{algorithmic}[1]
		\State {\bf Input} Number of sample paths $N$, number of iterations $M$, minibatch size $m$, discrete time points $t_{i}=\frac{i}{nT}\ (i=0,1,\cdots n)$
		\State {\bf Initialize} Generator parameters $\theta$, optimizer $Opt_{F}$, training dataset $\mathcal{D}_{\text{train}} = \varnothing$, training loss $L=\infty$
		
		\For{$i = 1, \ldots, N$}
		\State Sample initial points $\xi^{(i)}$
		\State Generate Brownian motion sample paths $B^{(i)}$
		\State Generate sample paths $X^{(i)}$
		\State Add input--output pairs $((\xi^{(i)},B^{(i)}), X^{(i)})$ to $\mathcal{D}_{\text{train}}$
		\EndFor
		
		\For{$j=1,\ldots, M$}
		\State Sample a minibatch $\mathcal{B} \subset \mathcal{D}_{\text{train}}$ of size $m$
		\State Generate sample paths $F_{\theta}(\xi^{(i)},B^{(i)}) $ for all $((\xi^{(i)},B^{(i)}), X^{(i)}) \in \mathcal{B}$
		\State Compute loss 
		\[L = \frac{1}{mn}\sum^{n}_{j=0} \sum_{\mathcal{B}} |X^{(i)}(t_{j})-F_{\theta}(\xi^{(i)},B^{(i)})(t_{j})|^{2}\]
		\State Update parameters $\theta$ via a call to $Opt_{F}(L,\theta)$
		\EndFor
	\end{algorithmic}
\end{algorithm}

\subsection{Simulation of path-dependent SDEs}

In this experiment, we train the MFNO to learn the solutions of two path-dependent SDEs of the form

\begin{equation}\label{eqn:SDE1}
dX^{1}(t)=(\alpha+\beta\int^{t}_{0} X^{1}(s)\, ds)\,dt + \sigma\, dB(t)
\end{equation}
and
\begin{equation}\label{eqn:SDE2}
dX^{2}(t)=\mu \,dt+ (\alpha+\beta\int^{t}_{0} X^{2}(s) \,ds)\,dB(t)\,.
\end{equation} 
We set the parameters to $\mu = 3.0$, $\alpha = 0.1$, $\beta = 0.03$, and $\sigma = 2.0$, with the initial condition drawn from a uniform distribution $U(0, 20)$.

\paragraph{Constructing Input--Output Pairs}  
As closed-form solutions for these SDEs are unavailable, we generate input--output pairs using the Euler scheme. First, we fix a time grid \( t_{0} < t_{1} < \cdots < t_{n} \) and simulate sample paths of the Brownian motion \( B \). For each instance, initial values \( \xi^{1} \) and \( \xi^{2}\) are drawn independently from \( U(0, 20) \). The numerical approximation for the first equation is given recursively by
\begin{align*}
X^{1}(t_{j+1}) &\approx X^{1}(t_{j}) + \big(\alpha + \beta \int_{0}^{t_{j}} X^{1}(s) \, ds\big) \Delta t_{j} + \sigma \Delta B(t_{j}) \\
&\approx X^{1}(t_{j}) + \big(\alpha + \beta \sum_{i=0}^{j} X^{1}(t_{i}) \Delta t_{i}\big) \Delta t_{j} + \sigma \Delta B(t_{j})
\end{align*}
for \( j = 0, 1, \ldots, n-1 \), where \( \Delta t_{j} = t_{j+1} - t_{j} \) and \( \Delta B(t_{j}) = B(t_{j+1}) - B(t_{j}) \sim N(0, \Delta t_{j}) \). Similarly, the second equation is approximated by
\begin{align*}
X^{2}(t_{j+1}) &\approx X^{2}(t_{j}) + \mu \Delta t_{j} + \big(\alpha + \beta \int_{0}^{t_{j}} X^{2}(s) \, ds\big) \Delta B(t_{j}) \\
&\approx X^{2}(t_{j}) + \mu \Delta t_{j} + \big(\alpha + \beta \sum_{i=0}^{j} X^{2}(t_{i}) \Delta t_{i}\big) \Delta B(t_{j})\,.
\end{align*}

\paragraph{Training}  
We construct a training dataset of 1,024 solution sample paths, each paired with its corresponding initial condition and driving Brownian motion path. The time grid is chosen as \( t_{0} = 0, t_{1} = 0.1, \ldots, t_{n} = 12.8 \), resulting in a grid size of 128 with a uniform step size \( \Delta t = 0.1 \). To ensure the training data accurately represents the true solution, we first simulate reference solutions using the Euler scheme with a much finer time step \( \Delta t = 0.1 \times 2^{-5} \). These high-fidelity solutions are then downsampled to the target resolution of 128. The MFNO architecture begins with mirror padding; it then lifts the input to a 32-channel latent space, and processes it through five Fourier layers, each with a width of 64. We use the Adam optimizer with a learning rate of \( 5 \times 10^{-4} \) and a weight decay of \( 3 \times 10^{-3} \). A StepLR scheduler with a step size of 100 and a decay factor \( \gamma = 0.9 \) is employed. The model is trained for 500 epochs with a batch size of 32. 

\paragraph{Testing}  
For evaluation, we generate separate test sets for each equation, each comprising 256 solution sample paths with their respective initial conditions and Brownian motion paths. These test solutions are computed using the Euler scheme with the finer time steps \( \Delta t\le 0.1 \times 2^{-5} \) and are subsequently resampled to various resolutions (128, 160, 192, 256, 320, 384, 512, 640, 832, and 1024) to assess the MFNO's ability to generalize across different discretizations.

\subsection{Simulation of fractional Brownian motion}

We demonstrate that the MFNO can be effectively trained to learn fBMs. In this experiment, we train the model to learn one-dimensional fBMs with Hurst parameters \( H = 0.25 \) and \( H = 0.75 \)  via regression.

\paragraph{Constructing Input--Output Pairs}  
We generate input--output pairs, comprising standard Brownian motion paths and their corresponding fBM paths, using the Cholesky decomposition method. The procedure is as follows:
\begin{enumerate}
	\item Select time points \( t_0 = 0, t_1, \ldots, t_n = T \) for sampling.
	\item Generate a vector \( Z = (z_1, z_2, \ldots, z_n) \), where each \( z_i \) is independently sampled from a normal distribution \( N(0, t_i) \).
	\item Compute the covariance matrix \( C \) defined by
	\[
	C(i, j) = \frac{1}{2}\big(|t_i|^{2H} + |t_j|^{2H} - |t_i - t_j|^{2H}\big), \quad i, j = 1, \ldots, n,
	\]
	where \( H \in (0, 1) \) is the Hurst parameter.
	\item Perform a Cholesky decomposition to find a lower triangular matrix \( A \) such that \( C = A^{\top} A \).
	\item The fBM sample points at times \( t_1, \ldots, t_n \) are given by the entries of \( AZ \), with the value at \( t_0 = 0 \) set to zero.  
\end{enumerate}

A key observation is that cumulatively summing the elements of \( Z \) yields a sample path of standard Brownian motion: explicitly, the values at \( t_0 = 0, t_1, \ldots, t_n \) are \( 0, z_1, z_1 + z_2, \ldots, z_1 + z_2 + \cdots + z_n \). Indeed, when \( H = 0.5 \), \( AZ \) reproduces this cumulative sum exactly.
This method provides a consistent framework for simultaneously generating paths of standard and fBM from the same underlying noise source, thereby enabling the construction of the required input--output pairs for training.

\paragraph{Training}  
Using this procedure, we construct a training dataset of 1,024 input--output pairs, each at a resolution of 128. We employ the same neural network architecture and optimization algorithm as in the path-dependent SDE experiments. We conduct separate experiments for \( H = 0.25 \) and \( H = 0.75 \). The MFNO architecture is the same as that employed for path-dependent SDEs.

\paragraph{Testing}  
For evaluation, we generate three test sets for each Hurst parameter, each containing 256 fBM sample paths at grid resolutions of 128, 160, 192, 256, 320, 384, 512, 640, 832, and 1024. These varied resolutions enable us to assess the MFNO’s generalization capability.

\subsection{Results: Comparison and Ablation Studies}

\paragraph{Comparative Analysis of Various Models}
We benchmark the MFNO against several baselines: the vanilla FNO, a zero-padded FNO (ZFNO), DeepONet, and two representative time-series models, TCN and LSTM. As DeepONet, TCN, and LSTM operate on a fixed temporal grid, we limit their evaluation to test data matching the training resolution; they are thus omitted from the variable-resolution experiments. To evaluate the impact of our padding strategy, we include both the vanilla FNO (no padding) and ZFNO (an FNO variant employing zero padding of the same size as MFNO's mirror padding) as control baselines. We also include DeepONet as a representative operator-learning baseline. For DeepONet, we configured the branch network with layers \([\,128,\,128,\,128,\,128]\) and the trunk network with layers \([\,128,\,128,\,128]\). The latent basis width was set to 300, and the model was trained with a learning rate of \(5 \times 10^{-4}\). For the TCN baseline, we adopted the architecture from \cite{Lea:16} with layer dimensions \([512,\,512,\,512,\,512,\,512,\,512,\,1]\). For the LSTM baseline, we used the standard configuration from \cite{HJ:97}, comprising two LSTM cells followed by a linear output map, with a hidden dimension of 512. The training and test datasets are identical to those described in Section~4.2. For TCN and LSTM, we set the learning rates to \(2 \times 10^{-4}\) and \(3 \times 10^{-4}\), respectively.

\begin{figure}[H]
	\centering
	\includegraphics[width=0.7\linewidth]{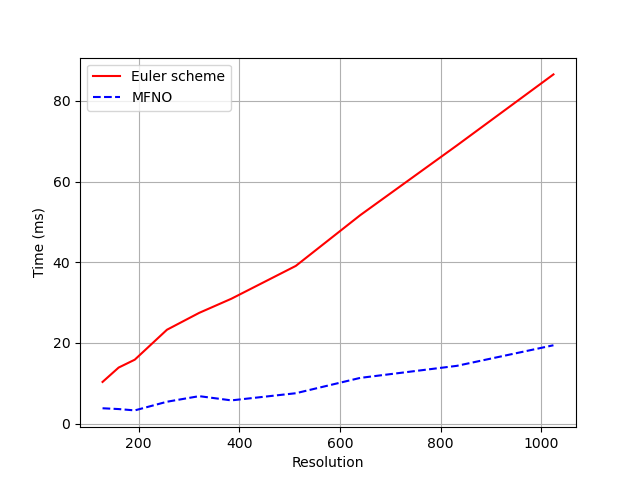}
	\caption{Comparison of inference times for path-dependent SDE \eqref{eqn:SDE1} across varying resolutions for the Euler–-Maruyama scheme and MFNO-based simulation. Reported values are the means over 100 independent runs.}
	\label{RT}
\end{figure}

Table~\ref{timepara} summarizes the number of parameters and inference speed for each model. Inference times were measured on test samples with a resolution of 256 under identical PyTorch GPU-parallel conditions. As shown, the FNO-based models achieve significantly faster inference than the traditional Euler solver. Given that the Euler scheme has a computational complexity of \(O(n^{2})\) while the FNO's complexity is \(O(n \log n)\) for an input resolution of \(n\), the MFNO becomes increasingly advantageous over the Euler method at higher resolutions, as demonstrated in Figure~\ref{RT}.

To evaluate test accuracy, we compute the relative \(l^{2}\) and relative \(l^{\infty}\) error norms. Table~\ref{l2_128} presents the average relative \(l^{2}\) norm for the resolution-128 test set, while Table~\ref{linf_128} shows the corresponding relative \(l^{\infty}\) norms. The results show that the FNO-based models deliver highly competitive performance across all tasks, with MFNO achieving the lowest error for the path-dependent SDE \eqref{eqn:SDE2}.

\begin{table}[H]
	\centering
	\resizebox{\columnwidth}{!}{%
		\begin{tabular}{lllll}
			\toprule
			\cmidrule(r){1-2}
			Model     & SDE 1 & SDE 2 &   fBM ($H=0.25$)   & fBM ($H=0.75$) \\
			\midrule
			MFNO & $\bf{(3.3\pm 1.1)\times10^{-4}}$  & $\bf{(7.4\pm3.5)\times10^{-5}}$   & $(1.3\pm0.16)\times10^{-2}$  &  $(1.4\pm0.21)\times10^{-2}$  \\
			ZFNO &  $(3.7\pm1.2)\times10^{-4}$ & $(8.1\pm3.6)\times10^{-5}$  &$(8.8\pm0.64)\times10^{-3}$   &  $(6.1\pm1.8)\times10^{-3}$  \\
			FNO & $(4.5\pm0.68)\times10^{-4}$ & $(8.6\pm3.6)\times10^{-5}$  & $\bf{(5.3\pm0.74)\times10^{-3}}$ &  $\bf{(3.7\pm0.39)\times10^{-3}}$   \\
			DeepOnet & $(1.4\pm0.12)\times10^{-2}$& $(3.2\pm0.12)\times10^{-3}$& $(2.8\pm0.13)\times10^{-1}$  & $(2.1\pm0.25)\times10{-2}$ \\
			TCN & $(2.1\pm0.62)\times10^{-3}$   & $(5.4\pm1.6)\times10^{-4}$ & $(1.5\pm0.21)\times10^{-2}\dagger$ &  $(1.0\pm0.19)\times10^{-2}$ \\
			LSTM & $(1.2\pm0.77)\times10^{-3}$ & $(1.8\pm1.6)\times10^{-4}$ &  $(5.4\pm1.3)\times10^{-3}$& $(4.5\pm2.0)\times10^{-3}$\\
			
			\bottomrule
		\end{tabular}
	}
	\caption{Average relative $l^{2}$ norm errors at resolution 128 across the two SDE tasks and fractional Brownian motion (fBM) cases with $H=0.25$ and $H=0.75$. Reported values are the mean $\pm$ standard deviation over 10 independent runs. For TCN on fBM with $H=0.25$, three runs exhibited unstable training with divergent errors; the statistics are computed from the remaining 7 runs.}
	\label{l2_128}
\end{table}

\begin{table}[H]
	\centering
	\resizebox{\columnwidth}{!}{%
		\begin{tabular}{lllll}
			\toprule
			\cmidrule(r){1-2}
			Model  & SDE 1 & SDE 2   &   fBM ($H=0.25$)   & fBM ($H=0.75$)\\
			\midrule
			MFNO & $(1.2\pm0.18)\times10^{-2}$  & $\bf{(1.3\pm0.22)\times10^{-2}}$   & $(9.1\pm0.61)\times10^{-2}$  &  $\bf{(2.4\pm0.34)\times10^{-2}}$  \\
			ZFNO &  $\bf{(1.1\pm0.22)}\times10^{-2}$ & $(1.4\pm0.15)\times10^{-2}$  &$(8.7\pm0.2)\times10^{-2}$   &  $(5.5\pm0.71)\times10^{-2}$  \\
			FNO &  $(1.5\pm0.25)\times10^{-2}$ &  $(1.5\pm0.26)\times 10^{-2}$ &  $\bf{(6.8\pm0.33)\times10^{-2}}$  & $(4.6\pm0.29)\times10^{-2}$\\
			DeepOnet &$(9.0\pm0.52)\times10^{-2}$ &$(1.2\pm0.021)\times10^{-1}$ & $(6.9\pm0.20)\times10^{-1}$  & $(1.7\pm0.10)\times10^{-1}$ \\
			TCN &  $(5.4\pm0.18)\times10^{-2}$  & $(4.7\pm0.47)\times10^{-2}$ & $(1.4\pm0.12)\times10^{-1}\dagger$& $(1.5\pm0.072)\times10^{-1}$\\
			LSTM &  $(2.0\pm0.34)\times10^{-2}$ & $(1.8\pm0.37)\times10^{-2}$ &$(1.6\pm0.029)\times10^{-1}$ &$(5.9\pm1.4)\times10^{-2}$\\
			
			\bottomrule
		\end{tabular}
	}
	\caption{Average relative $l^{\infty}$ norm errors at resolution 128 across the two SDE tasks and fractional Brownian motion (fBM) cases with $H=0.25$ and $H=0.75$. Reported values are the mean $\pm$ standard deviation over 10 independent runs. For TCN on fBM with $H=0.25$, three runs exhibited unstable training with divergent errors; the statistics are computed from the remaining 7 runs.}
	\label{linf_128}
\end{table}

\begin{table}[H]
	\centering
	\begin{tabular}{lll}
		\toprule
		\cmidrule(r){1-2}
		Model     &  Inference Time (ms)   & $\#$ of Parameters  \\
		\midrule
		Euler Method & $10.3$  &  \\
		MFNO & $2.3$ & $664,961$  \\
		ZFNO &  $2.0$ &  $''$  \\
		FNO &  $1.7$ &  $''$  \\
		DeepOnet & $0.63$  & $193,368$\\
		TCN & $27$ & $5,784,604$\\
		LSTM &$7.7$ & $3,158,529$ \\
		\bottomrule
	\end{tabular}
	\caption{Inference time for path-dependent SDE \eqref{eqn:SDE1} and number of parameters for each model. Reported values are the mean over 100 independent runs.}
	\label{timepara}
\end{table}

\begin{figure}[H]
	
	\begin{center}
		\includegraphics[width=0.49\textwidth]{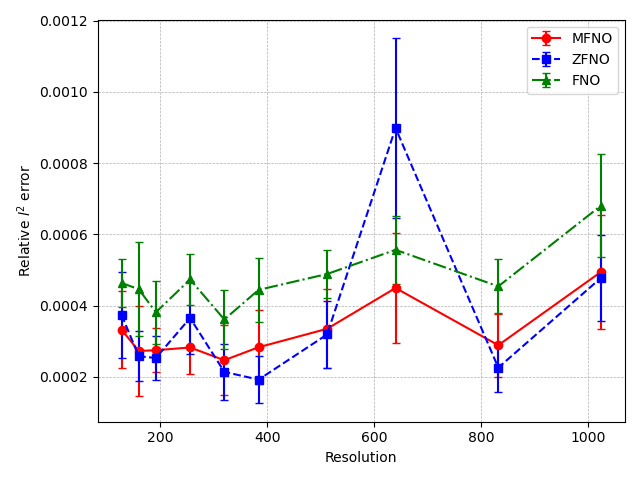}
		\includegraphics[width=0.49\textwidth]{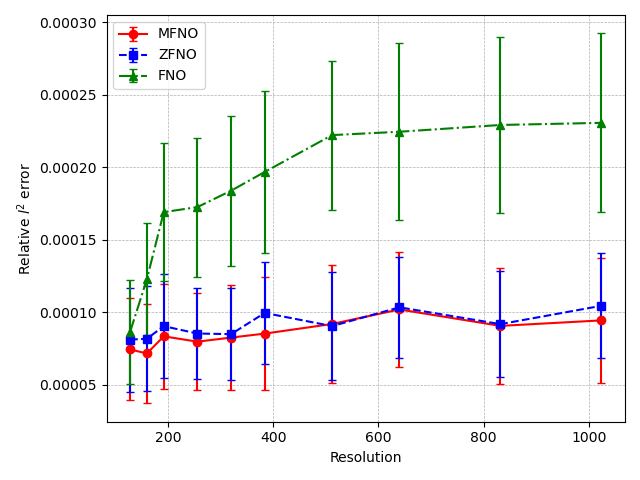}
		\\[\smallskipamount]
		\includegraphics[width=0.49\textwidth]{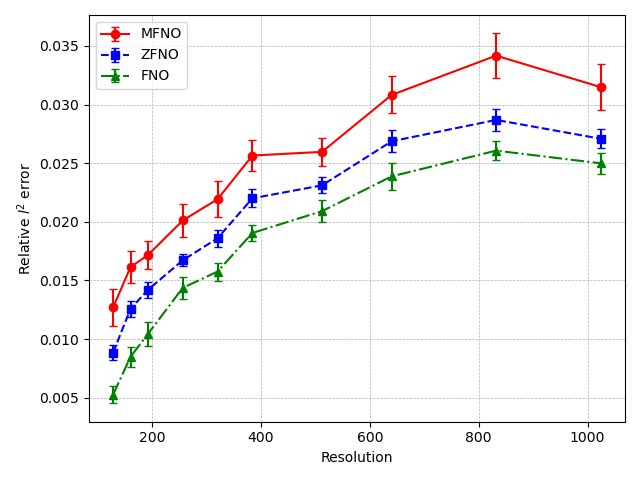}
		\includegraphics[width=0.49\textwidth]{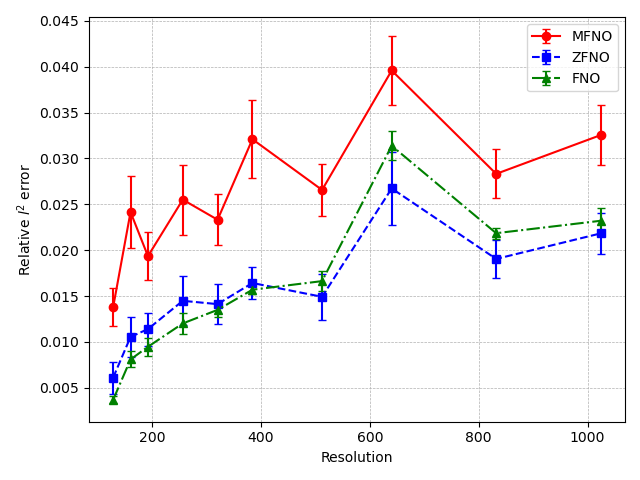}
	\end{center}
	\caption{Relative $l^{2}$ norm error trends with increasing test resolution for path-dependent SDEs and fBM: (top left) SDE \eqref{eqn:SDE1}, (top right) SDE \eqref{eqn:SDE2}, (bottom left) fBM $H=0.25$, and (bottom right) fBM $H=0.75$.}
	\label{resqual2}
\end{figure}

\begin{figure}[H]
	
	\begin{center}
		\includegraphics[width=0.49\textwidth]{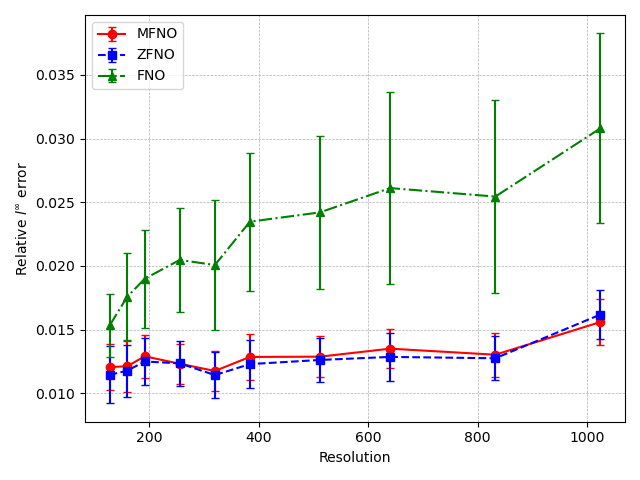}
		\includegraphics[width=0.49\textwidth]{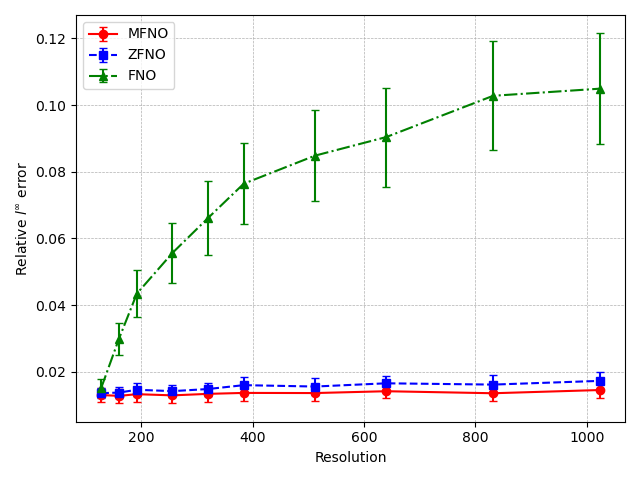}
		\\[\smallskipamount]
		\includegraphics[width=0.49\textwidth]{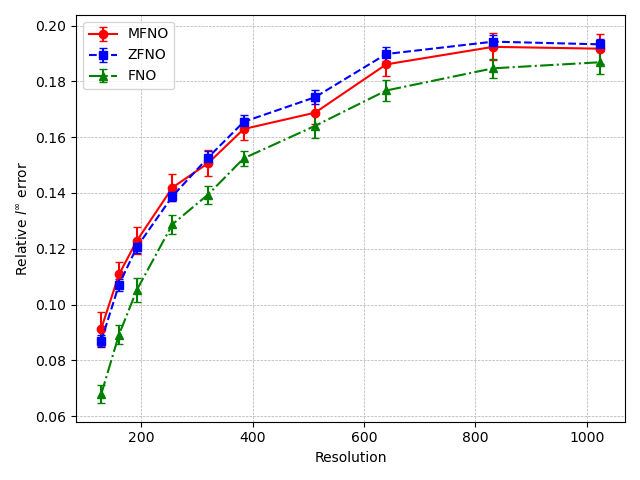}
		\includegraphics[width=0.49\textwidth]{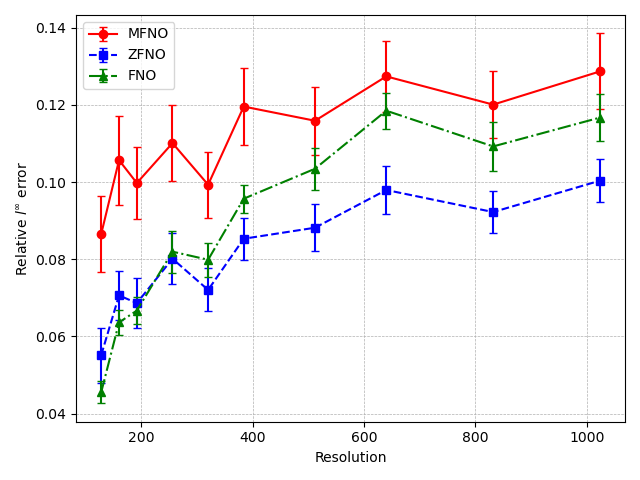}
	\end{center}
	\caption{Relative $l^{\infty}$ norm error trends with increasing test resolution for path-dependent SDEs and fBM: (top left) SDE \eqref{eqn:SDE1}, (top right) SDE \eqref{eqn:SDE2}, (bottom left) fBM $H=0.25$, and (bottom right) fBM $H=0.75$.}
	\label{resqualinf}
\end{figure}

\paragraph
{ Test Accuracies for Different Resolutions}
A key objective in designing MFNO was to enhance resolution generalization by addressing the boundary artifacts that arise in standard FNOs. To evaluate this property, we assessed the performance of models trained at a resolution of 128 on test samples of increasing resolution, specifically, 128, 160, 192, 256, 320, 384, 512, 640, 832, and 1024. The relative $l^{2}$ and $l^{\infty}$ errors for MFNO, ZFNO, and the vanilla FNO are reported in Figures~\ref{resqual2} and~\ref{resqualinf}.

Across the path-dependent SDE tasks, both MFNO and ZFNO exhibit strong resolution generalization, maintaining a nearly constant error as the grid is refined. In contrast, the vanilla FNO shows a consistent degradation in performance with increasing resolution. For fBM with Hurst parameter $H=0.25$, all models, including MFNO and ZFNO, demonstrate a significant loss of accuracy as resolution increases. This suggests that generalization is fundamentally constrained by the roughness of the underlying process. For the smoother fBM with $H=0.75$, MFNO achieves slightly better resolution stability than its counterparts, although it also starts with a higher error at the training resolution, rendering the overall advantage less clear.

These results indicate that our MFNO achieves performance comparable to ZFNO, a widely used baseline in empirical studies, on resolution generalization tasks. The vanilla FNO, by assuming periodicity of the input, suffers from wrap-around artifacts when applied to non-periodic signals, leading to instability under resolution refinement. In contrast, both MFNO and ZFNO extend the time domain to enforce periodicity, a critical requirement for the application of the Fourier transform in neural operator models. Despite this shared goal, the two approaches differ substantially in their theoretical properties. MFNO is explicitly designed to support rigorous mathematical analysis and is particularly amenable to proving approximation theorems. ZFNO, on the other hand, introduces artificial discontinuities at domain boundaries through zero-padding, making it challenging to analyze within a theoretical framework. Consequently, MFNO not only matches ZFNO in empirical performance but also offers significant advantages in terms of analytical tractability and theoretical rigor.

\section{Conclusion}
\label{sec:con}

In this work, we introduced the mirror-padded Fourier neural operator (MFNO), an architecture tailored for learning the solution operators of non-Markovian stochastic processes. We rigorously established approximation theorems demonstrating that the MFNO is capable of approximating solution operators for path-dependent SDEs and Lipschitz transformations of fBM.

To assess its practical effectiveness, we conducted extensive numerical experiments on both path-dependent SDEs and fBMs. Across these tasks, MFNO consistently achieved performance comparable or superior to that of baseline operator-learning models and conventional time-series methods in terms of both accuracy and computational efficiency. In particular, both MFNO and its zero-padded variant, ZFNO, demonstrated strong resolution generalization on the SDE tasks, whereas the vanilla FNO exhibited significant error degradation as resolution increased. For rougher processes, such as fBM with a low Hurst index, all FNO-based models showed reduced generalization, reflecting the inherent difficulty of the task rather than architectural limitations alone.

Overall, the MFNO provides a theoretically grounded and empirically robust framework for learning the solution operators of non-Markovian stochastic systems. Our results underscore the critical importance of boundary-aware architectural design in enhancing the stability and resolution adaptability of neural operator models.

$ $

\noindent{\bf{Acknowledgement}} Taeyoung Kim was supported by a KIAS Individual Grant (CG102201) and by the Center for Advanced Computation both at the Korea Institute for Advanced Study. 
Hyungbin Park was supported by the National Research Foundation of Korea (NRF) grants funded by the Ministry of Science and ICT (Nos.\ 2021R1C1C1011675 and 2022R1A5A6000840). 
Financial support from the Institute for Research in Finance and Economics of Seoul National University is gratefully acknowledged.

$ $

\noindent{\bf{Data Availability}} 
Data will be made available upon request.

\section*{Declarations}

\noindent{\bf{Conflict of interest}} 
The authors declare that they have no conflict of interest.

\bibliographystyle{spmpsci}
\bibliography{FNO}

\end{document}